\documentclass[letterpaper]{article} 
\usepackage{aaai2027}  
\usepackage[hyphens]{url}  
\usepackage{graphicx} 
\usepackage{natbib}  
\usepackage{caption} 
\usepackage{amsmath}
\usepackage{amssymb}
\usepackage{xcolor,colortbl}
\usepackage{color}
\usepackage{algorithm}
\usepackage{algorithmic}
\usepackage{booktabs}
\usepackage{tcolorbox}
\usepackage{xcolor}
\usepackage{amsmath}
\usepackage{graphicx}
\usepackage{amssymb}
\usepackage{pdfpages}

\usepackage{newfloat}
\usepackage{listings}
\DeclareCaptionStyle{ruled}{labelfont=normalfont,labelsep=colon,strut=off} 
\floatstyle{ruled}
\newfloat{listing}{tb}{lst}{}
\floatname{listing}{Listing}

\usepackage{booktabs}
\usepackage{multirow}
\makeatletter
\def\section{\@startsection {section}{1}{\z@}{-2.0ex plus -0.5ex minus -.2ex}{3pt plus 2pt minus 1pt}{\Large\bfseries\raggedright}}
\makeatother

\title{TRCA: Transition-wise Rubric Credit Assignment for Long-horizon LLM Agents}

\author{
    Huan Zhang\textsuperscript{\rm 1},
    Mingju Chen\textsuperscript{\rm 1},
    Dongxu Zhou\textsuperscript{\rm 1},
    Can Lv\textsuperscript{\rm 1},
    Heng Chang\textsuperscript{\rm 2},
    Sen Cui\textsuperscript{\rm 2},\\
    Faguo Wu\textsuperscript{\rm 1}\corresponding,
    Shiji Zhou\textsuperscript{\rm 1,\rm 2,\rm 3}\corresponding 
}

\affiliations{
    \textsuperscript{\rm 1}Beijing Advanced Innovation Center for Future Blockchain and Privacy Computing,
    School of Artificial Intelligence,\\ Beihang University,  \textsuperscript{\rm 2}Tsinghua University, 
    \textsuperscript{\rm 3}Beijing Academy of Artificial Intelligence (BAAI), \\[3pt]
    \textbf{Project Lead:} Heng Chang, 
    \textbf{Corresponding to:} Shiji Zhou
    \textless{}zhoushiji25@buaa.edu.cn\textgreater{}
}

\begin{document}

\maketitle

\begin{abstract}
Long-horizon large language model (LLM) agents are typically optimized with sparse terminal outcomes, providing coarse-grained credit assignment across multi-step interactions. Existing approaches either introduce pretrained process evaluators, incurring substantial annotation and inference costs, or derive step-level credit using successful trajectories as anchors. However, successful trajectories are extremely scarce during early-stage reinforcement learning in our case study, substantially weakening the effectiveness of anchor-based methods. To address this success-scarce credit assignment problem, we propose \textbf{T}ransition-wise \textbf{R}ubric \textbf{C}redit \textbf{A}ssignment (\textbf{TRCA}), which derives step-level supervision directly from action-induced transitions without learned evaluators or successful anchors. TRCA views interactive progress as requiring two complementary capabilities: acquiring task-relevant evidence from the environment and executing valid actions that meaningfully change it, while suppressing invalid, redundant, or regressive behavior. It therefore evaluates each transition using \textsc{Evidence}, \textsc{Execution}, and \textsc{Invalidity} rubrics. From the same rubric judgments, \textit{Foundational Rubric Reward} aggregates signed scores across the three categories to assess local transition quality, whereas \textit{Breakthrough Rubric Reward} tracks the cumulative coverage of positive \textsc{Evidence} and \textsc{Execution} conditions to reward newly established task progress. Combined with terminal outcomes, these signals yield fine-grained step-level advantages for policy optimization. Experiments on ALFWorld, WebShop, and seven search-augmented question-answering benchmarks show that TRCA consistently improves performance over the evaluated baselines. With Qwen2.5-7B-Instruct, TRCA improves the WebShop score by $6.0\%$--$12.6\%$ over competing baselines; with Qwen2.5-3B-Instruct, it improves the average SearchQA score by $1.9\%$--$18.3\%$. These results support the effectiveness of transition-wise rubric credit assignment for long-horizon interactive tasks in which successful anchors are scarce or absent. Code is available at https://github.com/1HUAN1/TRCA.

\end{abstract}

\section{Introduction}
\begin{figure}[t]
    \centering
    \vspace{-0.15cm}
    \includegraphics[width=\linewidth]{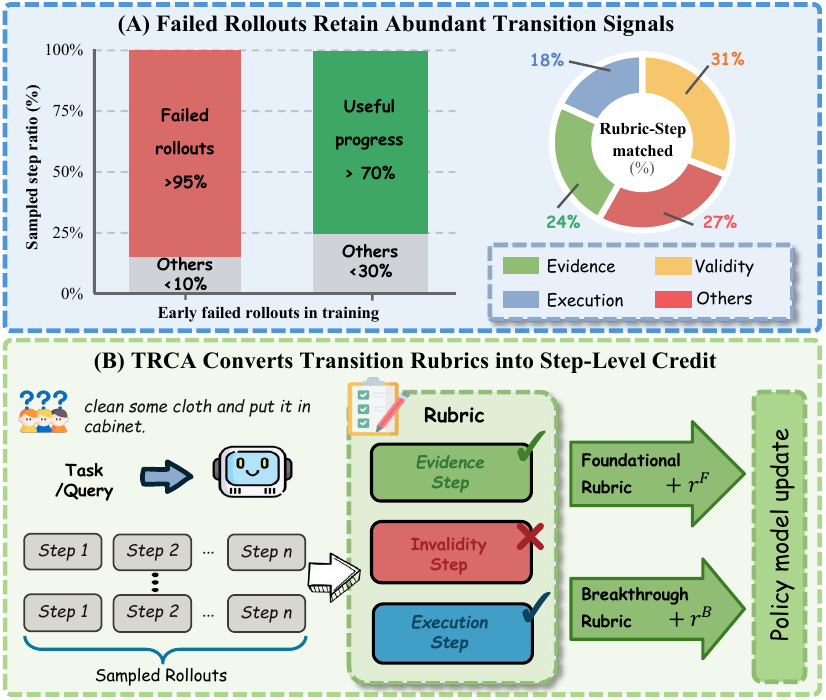}
    \caption{
\textbf{(A)} Failed rollouts dominate early training, yet over $70\%$ of their actions retain useful rubric-grounded transition signals.
\textbf{(B)} TRCA converts these step-level signals into foundational and breakthrough step credit without relying on successful anchors or external evaluators.
}
    \vspace{-0.3cm}
    \label{fig:Motivation}
\end{figure}

As large language model (LLM) agents become more capable, their applications are expanding from single-response generation to long-horizon interaction, including embodied planning~\citep{shridhar2021alfworld}, web navigation~\citep{NEURIPS2022_82ad13ec}, search-augmented reasoning, and multi-turn planning~\citep{pmlr-v235-xie24j,shao2026chinatravel}. However, when paired with sparse outcome rewards, PPO~\citep{schulman2017proximalpolicyoptimizationalgorithms}, GRPO~\citep{shao2024deepseekmathpushinglimitsmathematical}, and RLOO~\citep{ahmadian-etal-2024-back} provide only trajectory-level supervision for training on long-horizon tasks. Propagating the same outcome reward to all actions prevents them from distinguishing beneficial steps from harmful ones, obscuring the contribution of individual actions. This limitation motivates the use of fine-grained step-level supervision.

To address the above challenges, a natural solution is to introduce Process Reward Models (PRMs), which evaluate each intermediate step during rollouts to enable step-level credit assignment~\cite{yuan2025agentrtraininglanguagemodel}. However, existing process evaluation approaches based on supervised fine-tuning~\cite{10.1145/3774904.3792551} or step-by-step assessment with large language model judges typically require extensive human annotation or substantial additional inference. To avoid these labor and computational costs, another line of work derives step-level credit from structures constructed around successful trajectories ~\cite{NEURIPS2025_420c9f77,tan2026hindsightcreditassignmentlonghorizon,cheng2026trajectorylevelattributiongraphbasedcredit,feng2026rewardflowtopologyawarerewardpropagation}, thereby avoiding the overhead associated with step-wise evaluation. However, the effectiveness of these methods fundamentally depends on the availability of success anchors: reliable step-level credit propagation can only be established when sufficient successful trajectories are obtained. Consequently, their applicability is limited in exploration stages where successful trajectories are scarce.

To quantify the practical severity of this issue, we sample a group of rollouts from Qwen2.5-1.5B-Instruct on multiple long-horizon tasks. As shown in Figure~\ref{fig:Motivation} (A), 96.5\% of the sampled rollouts fail to achieve terminal success, leaving 85.6\% of task-conditioned training groups without any successful trajectory in early training. Under such a success-scarce regime, methods that rely on successful outcomes or success-conditioned structures to propagate rewards can provide only limited step-level credit. To examine whether failed rollouts are entirely uninformative, we further analyze their transitions using human annotations and a frontier LLM judge. Surprisingly, 72.2\% of the actions in failed trajectories still exhibit diagnostically useful transition signals, such as valid execution or measurable task progress. These observations suggest that reliable step-level credit remains accessible from environmental feedback even when terminal success is extremely sparse. Hence, these observations raise a core question: \textit{How can we assign reliable fine-grained credit without successful anchors or external evaluators?}

To answer this question, we propose \textbf{Transition-wise Rubric Credit Assignment (TRCA)}, a step-level credit assignment framework that derives supervision directly from action-induced transitions without relying on successful anchors or learned process evaluators. TRCA evaluates each transition using three rubrics: \textbf{\textsc{Evidence}} captures newly revealed task-relevant information, \textbf{\textsc{Invalidity}} identifies malformed or unexecutable actions, and \textbf{\textsc{Execution}} recognizes completed task-required operations or intermediate conditions. Based on these judgments, \textit{Foundational Rubric Reward} provides broad signed supervision, while \textit{Breakthrough Rubric Reward} assigns additional credit to transitions that newly cover previously unsatisfied task-relevant conditions. TRCA combines the resulting step-level rewards with the sparse trajectory outcome to construct fine-grained advantages for training. In this way, TRCA extracts reliable step-level credit from both successful and failed rollouts under success-scarce training conditions, as illustrated in Figure~\ref{fig:Motivation} (B). Our contributions are as follows:

\begin{itemize}
    \item We identify the success-scarce credit assignment problem in long-horizon agent training and quantify its severity through case studies. Our analysis shows that failed rollouts dominate early training, while their action-induced transitions still contain diagnostically useful signals.

    \item We propose \textbf{TRCA}, a transition-wise step-level credit assignment framework that evaluates action-induced transitions through Evidence, Invalidity, and Execution rubrics. TRCA constructs Foundational Rubric Reward for broad signed supervision and Breakthrough Rubric Reward for highlighting newly achieved task-relevant conditions, without successful anchors or external process evaluators.

    \item Extensive experiments on ALFWorld, WebShop, and seven search-augmented QA benchmarks demonstrate that TRCA consistently improves performance over the evaluated baselines across model scales. Ablation studies and hyperparameter analyses further validate the effectiveness and robustness of its components.
\end{itemize}

\begin{figure*}[t]
    \centering
    \includegraphics[width=\textwidth]{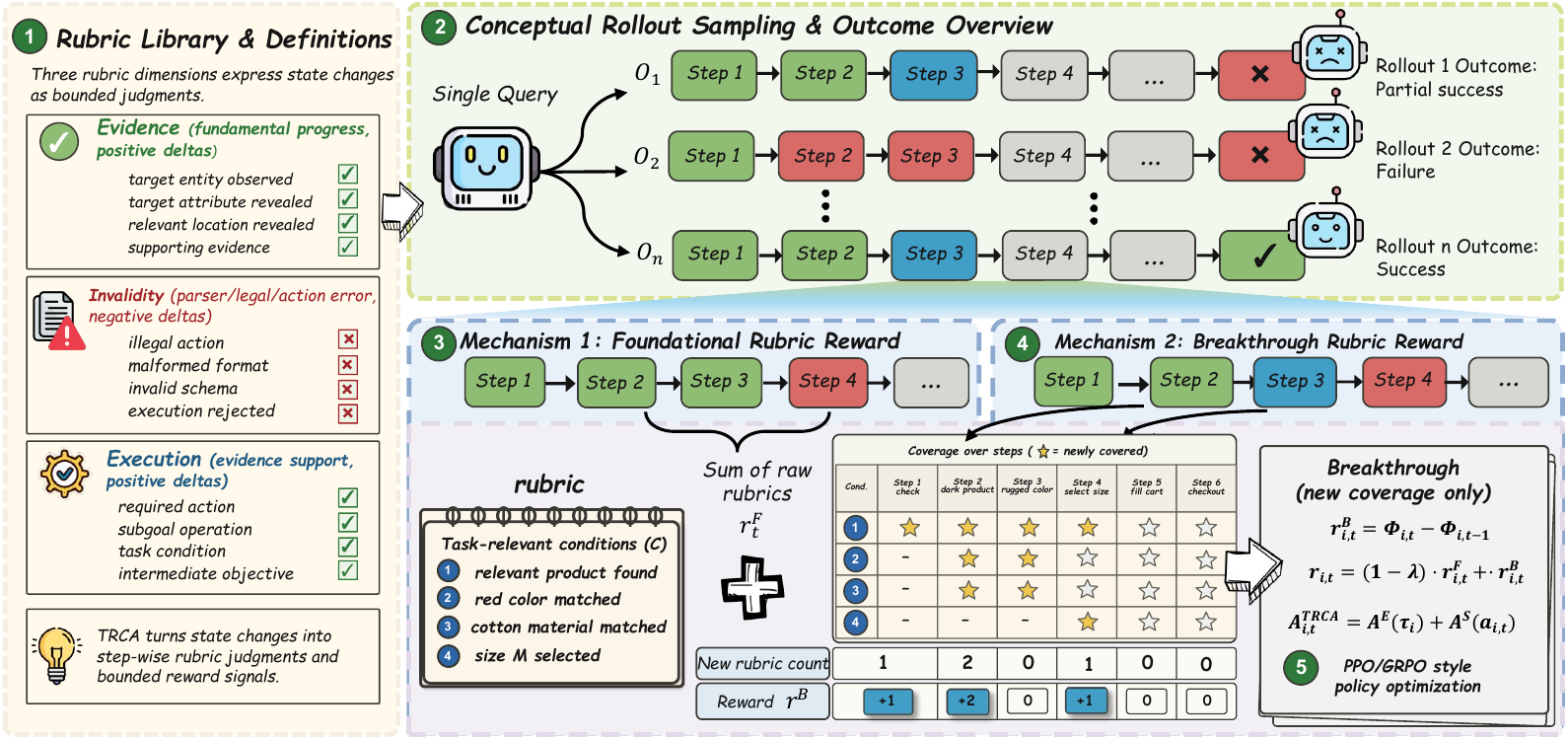}
    \caption{Overview of TRCA. Foundational Rubric Reward converts step-level rubric judgments into step-aligned credit, while breakthrough step rubric reward rewards newly covered task-relevant rubric conditions. The combined TRCA reward augments the original sparse environment reward to construct normalized step advantages, which are fused with the episode-relative trajectory advantage and optimized using the underlying PPO-style clipped objective.}
    \label{fig:trac_overview}
\end{figure*}
\section{Related Work}
\label{sec:related_work}

\subsection{Reinforcement Learning for LLM Agents}

Reinforcement learning for LLMs largely inherits its optimization machinery from RLHF. PPO stabilizes policy updates through a clipped surrogate objective and remains a widely used approach for LLM post-training~\citep{schulman2017proximalpolicyoptimizationalgorithms}. More recent methods simplify this actor-critic pipeline. GRPO removes the learned value function by estimating relative advantages within groups of sampled responses~\citep{shao2024deepseekmathpushinglimitsmathematical}, while RLOO employs leave-one-out baselines to obtain a simple critic-free policy-gradient estimator~\citep{ahmadian-etal-2024-back}. Although effective for single-response tasks, these methods provide limited guidance when a terminal outcome must be attributed to a long sequence of agent decisions. Recent work therefore extends group-relative optimization to multi-turn agents. GiGPO combines episode-level comparison with step-level grouping over recurrent anchor states, enabling finer-grained advantage estimation without additional rollouts~\citep{NEURIPS2025_420c9f77}. GraphGPO aggregates trajectories into a state-transition graph and estimates transition advantages using their structural relation to successful terminal states~\citep{cheng2026trajectorylevelattributiongraphbasedcredit}. These approaches substantially improve the granularity of advantage estimation by structure of success rollouts. 

\subsection{Process Supervision and Hindsight Credit Assignment}
LLM agents couple a policy model with an execution harness to support multi-step interaction with external environments, making fine-grained credit assignment over intermediate decisions a central challenge in agent optimization~\cite{chen2026harnessforgejointharnesspolicy,lv2026allmemagenticlifelongmemory,lin2026agenticharnessengineeringobservabilitydriven,chen2026amapreduceexecutingwidesearch}. Process supervision alleviates sparse feedback by assigning scores to intermediate steps. Human-supervised PRMs require large-scale step-level annotations, whereas automatic variants often rely on repeated Monte Carlo rollouts or tree search to construct training labels. Both introduce additional supervision or inference overhead~\citep{ICLR2024_aca97732,luo2024improvemathematicalreasoninglanguage}. AgentPRM extends this paradigm to interactive agent trajectories~\citep{10.1145/3774904.3792551}. Hindsight-based methods instead estimate earlier contributions after observing completed trajectories: HCAPO refines step-level values through hindsight reasoning, while AgentHER relabels failed or partially successful trajectories~\citep{tan2026hindsightcreditassignmentlonghorizon,ding2026agentherhindsightexperiencereplay}. These methods reduce explicit process evaluation, but may still depend on post-hoc goals, success proximity, or successful anchors that are scarce during early exploration.

\subsection{Rubric-Based Supervision and Reinforcement Learning}
Rubrics provide structured and interpretable supervision by decomposing complex or open-ended objectives into explicit evaluation criteria. Recent work has increasingly incorporated rubrics into LLM evaluation and post-training. Rubrics as Rewards (RaR) uses instance-specific rubric criteria as on-policy reward signals to extend reinforcement learning beyond domains with directly verifiable outcomes~\citep{gunjal2026rubrics}. GEAR further studies reward aggregation in rubric-based reinforcement learning by modeling prerequisite dependencies among rubric criteria, mitigating false credit propagation caused by treating interdependent criteria as independent reward signals~\citep{lv2026mitigatingfalsecreditpropagation}. Reinforcement Learning with Rubric Anchors further scales rubric-based rewards to open-ended tasks and uses structured criteria to provide fine-grained supervision over response quality~\citep{huang2025reinforcementlearningrubricanchors}. RuscaRL employs checklist-style rubrics both as guidance during rollout exploration and as references for reward computation, improving exploration in general reasoning tasks~\citep{zhou2026breakingexplorationbottleneckrubricscaffolded}. AdvancedIF and its RIFL pipeline similarly leverage expert-curated rubrics, rubric verification, and reward shaping to improve complex instruction following~\citep{he-etal-2026-advancedif}. Together, these studies demonstrate that rubrics can provide more explicit and decomposable supervision than holistic scalar feedback.



\section{Methodology}
\label{sec:method}

\subsection{Preliminaries}
\label{sec:method_preliminaries}

\subsubsection{Long-Horizon Agentic Tasks}
\label{sec:task_formulation}

Let \(x\sim p(\mathcal{X})\) denote a task instance and \(\pi_{\theta}\) an LLM policy. At interaction step \(t\), the agent observes state \(s_{i,t}\), which summarizes the current
observation, interaction history, and relevant environment feedback, and samples an action
\(a_{i,t}\sim\pi_{\theta}(\cdot\mid s_{i,t},x)\). After executing the action, the environment transitions to \(s_{i,t+1}\) and returns an environment reward \(r_{i,t}\).
A complete rollout is represented as the state-action trajectory$\tau_i=\{(x;s_{i,t},a_{i,t})\}_{t=1}^{T_i}$. For each task \(x\), the old policy \(\pi_{\theta_{\mathrm{old}}}\) samples \(N\) trajectories from the same initial condition. A long-horizon agentic task can be formulated as a partially observable Markov decision process, where the probability of trajectory \(\tau_i\) is
\begin{equation}
\pi_{\theta}\!\left(\tau_i\mid x,s_0\right) = \prod_{t=1}^{T_i} \pi_{\theta}\!\left( a_{i,t}\mid x,s_{i,\leq t} \right).
\label{eq:trajectory_probability}
\end{equation}
The environment assigns a binary terminal outcome
$R(\tau_i)=\sum_{t=1}^{T_i}r_{i,t}$, where \(R(\tau_i)=1\) indicates successful task completion and \(R(\tau_i)=0\) otherwise. Although this terminal outcome provides a global measure of rollout quality, it cannot reliably distinguish the precise contributions of individual actions within the entire trajectory.

\paragraph{Episode-Relative Advantage}
\label{sec:episode_relative_advantage}
Following the group-relative setting, the old policy samples a rollout group $G(x)=\{\tau_i\}_{i=1}^{N}$ for the same task and initial condition. We use the total return \(R(\tau_i)\) as a trajectory-level measure of task completion. The episode-relative advantage compares each trajectory with the other trajectories in the same group: \begin{equation}
A^{E}(\tau_i)
=\frac{R(\tau_i)-\mu\!\left(\{R(\tau_j)\}_{j=1}^{N}\right)}{\sigma\!\left(\{R(\tau_j)\}_{j=1}^{N}\right)}.
\label{eq:episode_relative_advantage}
\end{equation}
Thus, \(A^{E}(\tau_i)\) indicates whether rollout \(\tau_i\) performs better or worse than alternative rollouts sampled for the same task. When a rollout group contains no successful trajectories, we set $A^{E}(\tau_i)=0$. The step-level rewards become the primary source of learning signals for fine-grained credit assignment. 
\subsection{Transition-wise Rubric Design}
\label{sec:rubric_design}
To provide informative step-level reward when successful trajectories are sparse, we construct a rubric set $\mathcal R(x)$ that evaluates the state transition induced by each agent action. For a long-horizon task instance $x$ with rollout $\tau=(s_1,a_1,\ldots,s_T,a_T)$, the rubric is defined as \begin{equation} \mathcal{R}(x) = \left\{ \mathcal{R}^{\mathrm{Evidence}}(x), \mathcal{R}^{\mathrm{Invalidity}}(x), \mathcal{R}^{\mathrm{Execution}}(x) \right\}, \end{equation} where the three item types provide complementary credit signals for the action-induced transition $\xi_t=(s_t,a_t,s_{t+1})$. 
\textbf{\textsc{Evidence}} $\mathcal{R}^{\mathrm{Evidence}}(x)$ items identify task-relevant entities, attributes, locations, relations, or supporting facts revealed or supported by the current transition, providing positive credit for information acquisition that supports subsequent planning. 
\textbf{\textsc{Invalidity}} $\mathcal{R}^{\mathrm{Invalidity}}(x)$ items specify malformed, inadmissible, or otherwise unexecutable actions that cannot be correctly parsed or executed by the environment, contributing negative credit. \textbf{\textsc{Execution}} $\mathcal{R}^{\mathrm{Execution}}(x)$ items specify task requirements, subgoals, or intermediate objectives satisfied by the action-induced transition, providing additional positive credit for substantive task execution. This three-aspect schema separates information acquisition, invalid-action detection, and task-relevant execution. For each task type, all rubric items are instantiated offline by an LLM before training. No additional judge model is invoked during subsequent policy optimization. For a task instance $x$ belonging to task type $b$, we use the shared rubric library. Further details are provided in Appendix~\ref{app:benchmark_rubric_instantiation}.


\subsection{Transition-wise Rubric Rewards}
TRCA converts the step-level rubric judgments into two complementary reward signals: Foundational Rubric Reward and Breakthrough Rubric Reward.
\paragraph{Foundational Rubric Reward. }
\label{sec:Foundational}
Each judged rubric item is mapped to a signed item-level contribution according to its rubric category. Contributions associated with the same transition are then aggregated into a step-level reward and assigned only to the action tokens generated at the corresponding interaction step. We define the signed contribution of rubric item $j$ at transition (i,t) as follows:
\begin{equation}
q^i_{t,j} =
\begin{cases}
\displaystyle +r^{\mathrm{Evi}}/M_{\mathrm{Evi}}, & j \in \mathcal{R}^{Evi},\ \mathcal{O}_{i,t,j}=1, \\[4pt]
\displaystyle -\lvert r^{\mathrm{Inval}}\rvert/M_{\mathrm{Inval}}, & j\in \mathcal{R}^{Inval},\ \mathcal{O}_{i,t,j}=1, \\[4pt]
\displaystyle +r^{\mathrm{Exec}}/M_{\mathrm{Exec}}, & j \in \mathcal{R}^{Exec},\ \mathcal{O}_{i,t,j}=1, \\[4pt] 
0,& \text{otherwise}.
\end{cases}
\end{equation} Here, $q^i_{t,j}$ denotes the signed contribution of rubric item $j$ to transition and $M_c=|\mathcal{R}^{c}(x)|$ is the number of corresponding rubric items for task $x$. Dividing the budget by $M_c$ bounds the magnitude of each category's aggregate contribution regardless of rubric-list length. The Foundational Rubric Reward is then obtained by summing all item-level contributions at the same transition: $r^{\mathrm{F}}_{i,t} = \sum_{j\in\mathcal{R}(x)} q^i_{t,j}$, which aggregates both positive and negative judgments and is assigned only to the action tokens of the corresponding step. For each rubric item j, we define a binary semantic judgment \(\mathcal{O}_{i,t,j}\in\{0,1\}\), where \(\mathcal{O}_{i,t,j}=1\) if transition $(i,t)$ satisfies rubric item \(j\), and 0 otherwise. The judgment is computed from the observable semantic evidence associated with the transition. Further details are provided in Appendix~\ref{app:rule_based_evaluation}.

\paragraph{Breakthrough Rubric Reward.}
\label{sec:Breakthrough}

Foundational Rubric Reward evaluates each transition independently and may repeatedly reward actions that satisfy the same rubric, reducing the relative advantage of truly critical steps after normalization. We introduce Breakthrough Rubric Reward to address this issue by rewarding only newly satisfied rubric conditions. We define \(C_{i,t,j}=\max_{1\leq k\leq t}\mathcal{O}_{i,k,j}\), where \(\mathcal{O}_{i,t,j}=1\) if step \(t\) satisfies rubric item \(j\), and \(0\) otherwise. Breakthrough Rubric Reward then constructs a cumulative rubric potential from the covered positive rubric items as follows:
\begin{equation}
\Phi_{i,t} =\sum_{c\in\{\mathrm{Evi},\mathrm{Exec}\}}\frac{r^{c}}{M_{c}}\sum_{j\in\mathcal{R}^{c}(x)}
C_{i,t,j},
\label{eq:breakthrough_expanded}
\end{equation}
with $C_{i,0,j}=0$. And \(r^{c}>0\) denotes the reward budget assigned to category \(c\), and \(M_c=|\mathcal{R}^{c}(x)|\) is the number of rubric items in that category. We exclude Invalidity items from the cumulative potential because they describe transition-local errors and are already penalized by Foundational Rubric Reward. The Breakthrough Rubric Reward assigned to transition \(t\) is defined as the temporal difference of the cumulative potential: \(r^{\mathrm{B}}_{i,t}=\Phi_{i,t}-\Phi_{i,t-1}\). Equivalently, the temporal difference can be expanded as
\begin{equation}
r^{\mathrm{B}}_{i,t} = \sum_{c\in\{\mathrm{Evi},\mathrm{Exec}\}} \frac{r^{c}}{M_{c}} \sum_{j\in\mathcal{R}^{c}(x)} \left(1-C_{i,t-1,j}\right) \mathcal{O}_{i,t,j}.
\label{eq:kpsc_expanded}
\end{equation}
Consequently, rubric item \(j\) contributes to Breakthrough Rubric Reward only when it is satisfied by the current step, \(\mathcal{O}_{i,t,j}=1\), and has not been covered by any earlier step, \(C_{i,t-1,j}=0\). Overall, TRCA consists of two complementary reward constructions: Foundational Rubric Reward, which evaluates the direct rubric evidence triggered by each action, and Breakthrough Rubric Reward, which measures transition-induced increases in cumulative task-relevant rubric coverage. The two rewards are combined as 
\begin{equation}
r_{i,t}^{\mathrm{TRCA}} = (1-\lambda)\cdot r_{i,t}^{\mathrm{F}} + \lambda\cdot r_{i,t}^{\mathrm{B}} ,
\label{eq:trca_step_reward}
\end{equation}
Here, \(\lambda\) $\in[0,1]$ controls the contribution of Breakthrough Rubric Reward relative to Foundational Rubric Reward. Importantly, TRCA constructs step-level rewards without requiring successful anchor states.

\begin{table*}[t]
\centering
\setlength{\tabcolsep}{5pt} %
\begin{tabular}{ll cccccc cc}
\toprule
\multirow{2}{*}{\textbf{Type}} 
& \multirow{2}{*}{\textbf{Method}} 
& \multicolumn{6}{c}{\textbf{ALFWorld}} 
& \multicolumn{2}{c}{\textbf{WebShop}} \\
\cmidrule(lr){3-8} \cmidrule(lr){9-10}
& & \textbf{Clean} & \textbf{Pick} & \textbf{Cool} & \textbf{Heat} 
& \textbf{Pick2} & \textbf{All} & \textbf{Score} & \textbf{Succ.} \\
\midrule
\multicolumn{10}{l}{\textit{Closed-Source Model}} \\
\midrule
\multirow{2}{*}{Prompting} 
& GPT-4o 
& 31.2 & 75.3 & 21.6 & 56.7 & 49.8 & 48.0 & 31.8 & 23.7 \\
& Gemini-2.5-Pro 
& 62.1 & 92.8 & 26.6 & 69.0 & 58.7 & 60.3 & 42.5 & 35.9 \\
\midrule
\multicolumn{10}{l}{\textit{Qwen2.5-1.5B-Instruct}} \\
\midrule
\multirow{3}{*}{Prompting} 
& Qwen2.5 
& 3.3 & 5.9 & 4.2 & 9.7 & 0.0 & 4.1 & 23.1 & 5.2 \\
& ReAct 
& 15.7 & 17.4 & 7.7 & 6.2 & 2.0 & 12.8 & 40.1 & 11.3 \\
& Reflexion 
& 21.7 & 35.3 & 19.4 & 13.6 & 3.7 & 21.8 & 55.8 & 21.9 \\
\cmidrule(lr){2-10}
\multirow{7}{*}{RL Training} 
& PPO
& 57.1$_{(4.9)}$ & 64.8$_{(3.5)}$ & 46.4$_{(4.0)}$ 
& 60.6$_{(6.6)}$ & 47.4$_{(1.9)}$ & 54.4$_{(3.1)}$ 
& 73.8$_{(3.0)}$ & 51.5$_{(2.9)}$ \\
& RLOO
& 71.0$_{(5.9)}$ & 88.3$_{(3.0)}$ & 66.4$_{(5.5)}$ 
& 62.8$_{(8.7)}$ & 56.9$_{(4.7)}$ & 69.7$_{(2.5)}$ 
& 73.9$_{(5.6)}$ & 52.1$_{(6.7)}$ \\
& GRPO
& 73.9$_{(6.8)}$ & 82.9$_{(3.6)}$ & 77.8$_{(4.5)}$ 
& 78.6$_{(0.0)}$ & 71.4$_{(3.9)}$ & 77.9$_{(1.3)}$ 
& 84.7$_{(0.5)}$ & 71.4$_{(2.1)}$ \\
& GiGPO 
& 94.8$_{(3.8)}$ & 94.4$_{(5.9)}$ & 79.8$_{(4.7)}$ 
& 94.4$_{(7.8)}$ & 76.4$_{(5.4)}$ & 86.7$_{(1.7)}$ 
& 83.1$_{(1.6)}$ & 65.0$_{(3.2)}$ \\
& HCAPO
& \textbf{97.6$_{(1.8)}$} & 88.6$_{(7.0)}$ & 84.2$_{(0.0)}$ 
& 90.7$_{(6.9)}$ & 74.2$_{(6.9)}$ & 87.0$_{(4.1)}$ 
& 83.8$_{(0.7)}$ & 68.5$_{(1.0)}$ \\
& GraphGPO
& 88.2$_{(5.0)}$ & \textbf{97.8$_{(3.2)}$} & 91.3$_{(1.8)}$ 
& 80.2$_{(11.1)}$ & 86.6$_{(2.6)}$ & 91.7$_{(0.4)}$ 
& 86.7$_{(1.6)}$ & 75.5$_{(1.2)}$ \\
& \textbf{Ours}
& 92.5$_{(4.3)}$ & 95.9$_{(2.2)}$ & \textbf{96.6$_{(1.5)}$} 
& \textbf{97.8$_{(0.9)}$} & \textbf{86.8$_{(6.9)}$} 
& \textbf{92.0$_{(2.7)}$} & \textbf{90.5$_{(0.9)}$} 
& \textbf{78.4$_{(1.6)}$} \\
\midrule
\multicolumn{10}{l}{\textit{Qwen2.5-7B-Instruct}} \\
\midrule
\multirow{3}{*}{Prompting} 
& Qwen2.5 
& 19.3 & 33.4 & 2.8 & 6.9 & 3.2 & 14.8 & 26.4 & 7.8 \\
& ReAct 
& 34.3 & 48.5 & 18.2 & 13.2 & 17.6 & 31.2 & 46.2 & 19.5 \\
& Reflexion 
& 44.9 & 62.0 & 36.3 & 30.9 & 23.8 & 42.7 & 58.1 & 28.8 \\
\cmidrule(lr){2-10}
\multirow{7}{*}{RL Training} 
& PPO
& 92.5$_{(2.4)}$ & 92.3$_{(4.0)}$ & 80.3$_{(2.0)}$ 
& 89.5$_{(7.0)}$ & 68.8$_{(8.3)}$ & 80.4$_{(2.7)}$ 
& 81.4$_{(3.1)}$ & 68.7$_{(5.1)}$ \\
& RLOO
& 87.3$_{(5.8)}$ & 87.6$_{(4.3)}$ & 71.9$_{(5.2)}$ 
& 81.3$_{(7.6)}$ & 48.9$_{(8.4)}$ & 75.5$_{(4.6)}$ 
& 80.3$_{(3.2)}$ & 65.7$_{(4.0)}$ \\
& GRPO
& 77.9$_{(4.6)}$ & 89.0$_{(5.3)}$ & 90.7$_{(5.2)}$ 
& 78.6$_{(0.0)}$ & 71.4$_{(3.9)}$ & 83.3$_{(2.1)}$ 
& 84.3$_{(1.3)}$ & 75.0$_{(2.8)}$ \\
& GiGPO 
& \textbf{98.8$_{(1.6)}$} & 97.7$_{(1.6)}$ & 89.3$_{(8.2)}$ 
& 83.7$_{(7.2)}$ & 79.2$_{(6.6)}$ & 90.8$_{(1.3)}$ 
& 84.4$_{(2.9)}$ & 72.8$_{(3.2)}$ \\
& HCAPO
& 97.3$_{(1.9)}$ & \textbf{99.1$_{(1.3)}$} & 90.8$_{(6.6)}$ 
& 81.8$_{(8.8)}$ & 81.9$_{(10.0)}$ & 91.4$_{(2.3)}$ 
& 85.1$_{(1.3)}$ & 73.8$_{(2.8)}$ \\
& GraphGPO
& 89.6$_{(1.7)}$ & 98.0$_{(1.0)}$ & 92.5$_{(1.1)}$ 
& 90.9$_{(5.3)}$ & 90.4$_{(2.4)}$ & 93.3$_{(1.2)}$ 
& 86.9$_{(0.7)}$ & 80.3$_{(2.3)}$ \\
& \textbf{Ours}
& 95.0$_{(2.1)}$ & 97.7$_{(1.2)}$ & \textbf{100.0$_{(0.0)}$} 
& \textbf{93.3$_{(3.1)}$} & \textbf{94.1$_{(5.8)}$} 
& \textbf{94.5$_{(1.4)}$} & \textbf{92.9$_{(1.9)}$} 
& \textbf{83.8$_{(1.6)}$} \\
\bottomrule
\end{tabular}

\caption{Performance on ALFWorld and WebShop. Results are averaged over 3 random seeds. For ALFWorld, we report the average success rate (\%) for each subtask as well as the overall result. For WebShop, we report both the average score and the average success rate (\%). We compare TRCA with other representative baselines. Best results are \textbf{bolded}.}
\label{tab:main_results}
\vspace{-0.3cm}
\end{table*}

\subsection{Policy Optimization}
\label{sec:policy_optimization}

Following the group-in-group framework, steps encountered under similar decision contexts are clustered according to environment-specific state information. For each similar state \(\tilde{s}\), we define the corresponding step-level group as
\begin{equation}
G^{S}(\tilde{s})=\left\{(i,t)\;\middle|\;s_{i,t}\simeq\tilde{s},\ 1\leq i\leq N,\ 1\leq t\leq T_i\right\},
\label{eq:step_group}
\end{equation}
where \(s_{i,t}\simeq\tilde{s}\) indicates that \(s_{i,t}\) belongs to the decision context represented by \(\tilde{s}\). Rather than requiring exact textual matches, the grouping identifies semantically equivalent or structurally consistent states using environment-specific information, such as page type, product identity, selected options, task predicates, or interaction stage.

The original environment reward \(r_{i,t}\) is sparse and is typically assigned only at the terminal transition, with \(r_{i,t}=0\) for \(t<T_i\) and \(r_{i,T_i}=R(\tau_i)\). TRCA combines the sparse environment reward with the step-level rubric reward to compute the completion-aware return:
\begin{equation}
R_{i,t}=\sum_{k=t}^{T_i}\gamma^{k-t}\left(r_{i,k}+r_{i,k}^{\mathrm{TRCA}}\right),
\label{eq:completion_aware_return}
\end{equation}
where \(\gamma\in[0,1]\) is the discount factor. Analogous to the episode-relative advantage, the step-relative advantage is computed by normalizing the completion-aware return within the corresponding step-level group:
\begin{equation}
A^{S}(a_{i,t})=
\frac{R_{i,t}-\mu\!\left(\left\{R_{j,k}\mid(j,k)\in G^{S}(\tilde{s})\right\}\right)}
{\sigma\!\left(\left\{R_{j,k}\mid(j,k)\in G^{S}(\tilde{s})\right\}\right)},
\label{eq:step_relative_advantage}
\end{equation}
Here, the mean and standard deviation are computed over the completion-aware returns collected from the same comparable decision context. While \(A^{E}(\tau_i)\) evaluates the relative quality of the complete rollout, \(A^{S}(a_{i,t})\) estimates the long-term utility of the individual action. 
TRCA combines the episode-relative and step-relative signals to construct the final advantage: $A_{i,t}^{\mathrm{TRCA}} = A^{E}(\tau_i) + A^{S}(a_{i,t}),$ the clipped policy optimization objective of TRCA is:
\begin{equation}
\label{eq:final_objective} 
\begin{aligned}
  &\mathcal{J}_{\mathrm{TRCA}}(\theta) =\mathbb{E}
   \Bigl[\frac{1}{N}
    \sum_{i=1}^{N}\frac{1}{T_i}
    \sum_{t=1}^{T_i}
    \min \Big( \rho_{i,t}(\theta)A_{i,t}^{\text{TRCA}}, \\
    &\text{clip}(\rho_{i,t}(\theta), 1-\epsilon, 1+\epsilon) A_{i,t}^{\text{TRCA}} \Big) \Bigl] - \beta_{\text{KL}} \mathbb{D}_{\text{KL}}(\pi_\theta || \pi_{\text{ref}})
\end{aligned}
\end{equation}
Where $\rho_{i,t}(\theta) = \frac{\pi_\theta( a_{i,t} \mid \ {s}_{i,t}, x)}{\pi_{\theta_{\text{old}}}(a_{i,t} \mid \ {s}_{i,t}, x)}$ is the action-level importance ratio and \(\beta_{\mathrm{KL}}\) controls regularization toward the reference policy. The KL term is omitted when explicit KL regularization is not used. The complete training procedure is provided in Appendix~\ref{app:training_procedure}.


\begin{table*}[t]
\centering
\setlength{\tabcolsep}{5pt} %
\begin{tabular}{ll ccc cccc c}
\toprule
\multirow{2}{*}{\textbf{Type}} & \multirow{2}{*}{\textbf{Method}} & \multicolumn{3}{c}{\textbf{Single-Hop QA}} & \multicolumn{4}{c}{\textbf{Multi-Hop QA}} & \multirow{2}{*}{\textbf{Avg.}} \\
\cmidrule(lr){3-5} \cmidrule(lr){6-9}
& & \textbf{NQ}$\dagger$ & \textbf{TriviaQA}$\star$ & \textbf{PopQA}$\star$ & \textbf{HotpotQA}$\dagger$ & \textbf{2Wiki}$\star$ & \textbf{MuSiQue}$\star$ & \textbf{Bamboogle}$\star$ & \\
\midrule
\multicolumn{10}{l}{\textit{Base Model: Qwen2.5-3B-Instruct}} \\
\midrule
\multirow{7}{*}{RL Training} 
 & R1-Instruct & 27.0 & 53.7 & 19.9 & 23.7 & 29.2 & 7.2 & 29.3 & 27.1 \\
 & Search-R1 & 34.1 & 54.5 & 37.8 & 32.4 & 31.9 & 10.3 & 26.4 & 32.5 \\
 & ZeroSearch & 41.4 & 57.4 & 44.8 & 27.4 & 30.0 & 9.8 & 11.1 & 31.7 \\
 & StepSearch & -- & -- & -- & 34.5 & 32.0 & 17.4  & -- & 34.4 \\
 & GiGPO & 42.0 & 59.5 & 42.4 & 36.9 & 37.0 & 12.6 & 64.1 & 42.1 \\
 & HCAPO  &44.4 & 60.5 &45.5& 38.6 & 36.3 & 14.8 & 64.5 & 43.5 \\
& IGPO & 36.4 & 55.8 & 36.3 & 32.8 & 31.6 & 10.6 &56.8 & 37.2 \\
\cmidrule(lr){2-10}
RL Training & \textbf{TRCA} &\textbf{46.7} &\textbf{62.4}  &\textbf{45.7}  & \textbf{40.5}& \textbf{40.0} &\textbf{18.0} & \textbf{64.8}  & \textbf{45.4} \\
\midrule
\multicolumn{10}{l}{\textit{Base Model: Qwen2.5-7B-Instruct}} \\
\midrule
\multirow{7}{*}{RL Training} 
 & R1-Instruct & 21.0 & 44.9 & 17.1 & 20.8 & 27.5 & 6.0 & 19.2 & 22.4 \\
 & Search-R1 & 39.3 & 61.0 & 39.7 & 37.0 & 40.1 & 14.6 & 36.8 & 38.5 \\
 & ZeroSearch & 43.6 & 61.8 & \textbf{51.5} & 34.6 & 35.2 & 18.4 & 27.8 & 39.1 \\
 & StepSearch & -- & -- & -- & 38.6 & 36.6 & \textbf{22.6} & -- & 40.0 \\
 & GiGPO & 46.4 & 64.7 & 46.1 & 41.6 & 43.6 & 18.9& 68.9 & 47.2\\
 & HCAPO & 46.1 & 65.5 & 47.6 & 42.1 & 43.1 & 17.7 & 69.0 &47.3  \\
& IGPO & 46.5 & 64.1 &47.0 & 42.0& 41.7 & 18.5 & 67.2&46.7 \\
\cmidrule(lr){2-10}
 RL Training & \textbf{TRCA} & \textbf{47.9}&\textbf{65.8}  & 48.2 &\textbf{42.3} &\textbf{43.6}  &20.9  & \textbf{70.4} & \textbf{48.4 }\\
\bottomrule
\end{tabular}
\caption{Performance on search-augmented QA tasks. $\dagger$ and $\star$ indicate in-domain and out-of-domain datasets, respectively. Bold indicates the best performance in each category.}
\vspace{-0.1cm}
\label{tab:qa_results}
\end{table*}

\section{Experiments}
\label{sec:experiments}
We evaluate TRCA on two long-horizon interactive environments and seven search-augmented question-answering benchmarks. Our experiments examine (1) the overall effectiveness of TRCA across task domains and model scales, (2) its performance on multi-turn search tasks, (3) the contributions of Foundational Rubric Reward and Breakthrough Rubric Reward, and (4) its sensitivity to the mixing coefficient \(\lambda\) between the two reward components.

\subsection{Experimental Setup}
\label{sec:experimental_setup}

\paragraph{Benchmarks and metrics.}
We first evaluate TRCA on ALFWorld~\cite{shridhar2021alfworld} and WebShop~\cite{NEURIPS2022_82ad13ec}. ALFWorld is a text-based embodied environment in which an agent must complete household tasks through multi-step interactions. We report the success rate for each category and the overall success rate. WebShop is an interactive shopping environment in which an agent navigates a simulated e-commerce website to identify and purchase a product that satisfies a natural-language instruction. Following prior work, we report both the average task score, which reflects partial attribute satisfaction, and the success rate, which measures exact task completion.
We further evaluate TRCA on search-augmented QA, including three single-hop datasets---Natural Questions (NQ)~\cite{kwiatkowski-etal-2019-natural}, TriviaQA~\cite{joshi-etal-2017-triviaqa}, and PopQA~\cite{mallen-etal-2023-trust}---and four multi-hop datasets---HotpotQA~\cite{yang-etal-2018-hotpotqa}, 2WikiMultiHopQA (2Wiki)~\cite{ho-etal-2020-constructing}, MuSiQue~\cite{trivedi2022musiquemultihopquestionssinglehop}, and Bamboogle~\cite{press-etal-2023-measuring}. We treat NQ and HotpotQA as in-domain benchmarks and use the remaining datasets to assess out-of-domain generalization. We report task success rate under strict binary normalized Exact Match (EM), counting an example as successful only when the normalized final answer exactly matches a ground-truth alias.
\vspace{-0.1cm}
\paragraph{Baselines.}
For LLMs, we compare TRCA with several competitive baselines, including: 
1) Closed-source LLMs: GPT-4o and Gemini-2.5-Pro~\cite{article}. 
2) Prompting-based agents: ReAct~\cite{yao2023reactsynergizingreasoningacting}, and Reflexion~\cite{shinn2023reflexionlanguageagentsverbal}, which are instantiated with the Qwen2.5 base policy ~\cite{qwen2025qwen25technicalreport}.
3) RL-based training methods: We choose models as follows, the actor-critic method PPO~\cite{schulman2017proximalpolicyoptimizationalgorithms}, the critic-free methods RLOO~\cite{kool2019buy,ahmadian-etal-2024-back} and GRPO~\cite{shao2024deepseekmathpushinglimitsmathematical}; and recent step-level credit-assignment methods GiGPO~\cite{NEURIPS2025_420c9f77}, HCAPO~\cite{tan2026hindsightcreditassignmentlonghorizon}, and GraphGPO~\cite{cheng2026trajectorylevelattributiongraphbasedcredit}. For \textbf{search-augmented QA} tasks, following the experimental protocol in prior work ~\citep{NEURIPS2025_420c9f77}, we compare with R1-Instruct, Search-R1~\cite{jin2025searchr1trainingllmsreason}, ZeroSearch~\cite{sun2026zerosearchincentivizesearchcapability}, StepSearch~\cite{wang2025stepsearchignitingllmssearch}, GiGPO, HCAPO, and IGPO~\cite{wang2026informationgainbasedpolicyoptimization}. 

\paragraph{Implementation details.}
For ALFWorld and WebShop, we use Qwen2.5-1.5B-Instruct and Qwen2.5-7B-Instruct~\cite{qwen2025qwen25technicalreport} as the policy models. All group-based methods use a rollout group size of 8. The agent generates its reasoning within \texttt{<think>} tags and its executable action within \texttt{<action>} tags. Unless otherwise specified, we set the mixing coefficient between Foundational Rubric Reward and Breakthrough Rubric Reward to \(\lambda=0.8\). Results on ALFWorld and WebShop are averaged over three random seeds. For search-augmented QA, we use Qwen2.5-3B-Instruct and Qwen2.5-7B-Instruct, while keeping the training and evaluation settings identical to those of GiGPO. Full details are provided in Appendix~\ref{app:experimental_details}.

\begin{figure*}[t]
    \centering
    \includegraphics[width=\textwidth, height=3.5cm]{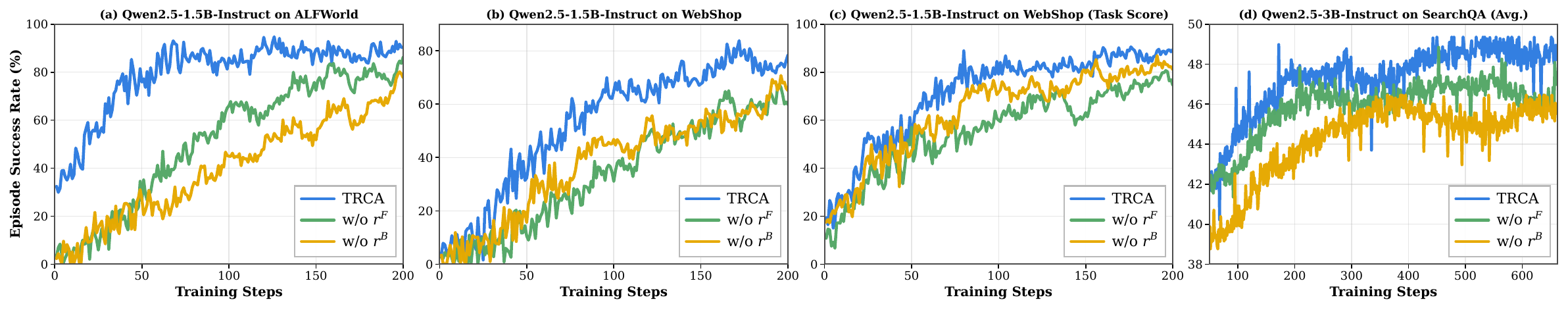}
    \caption{Training dynamics of TRCA and its ablated variants on ALFWorld, WebShop, and SearchQA. Removing either Foundational Rubric Reward ($r^{\mathrm{F}}$) or Breakthrough Rubric Reward ($r^{\mathrm{B}}$) consistently degrades performance.}
    \label{fig:trca_ablation_1x4_front50_noisy}

\end{figure*}
\vspace{-0.1cm}
\begin{figure}[h]
    \centering
    \includegraphics[width=\linewidth]{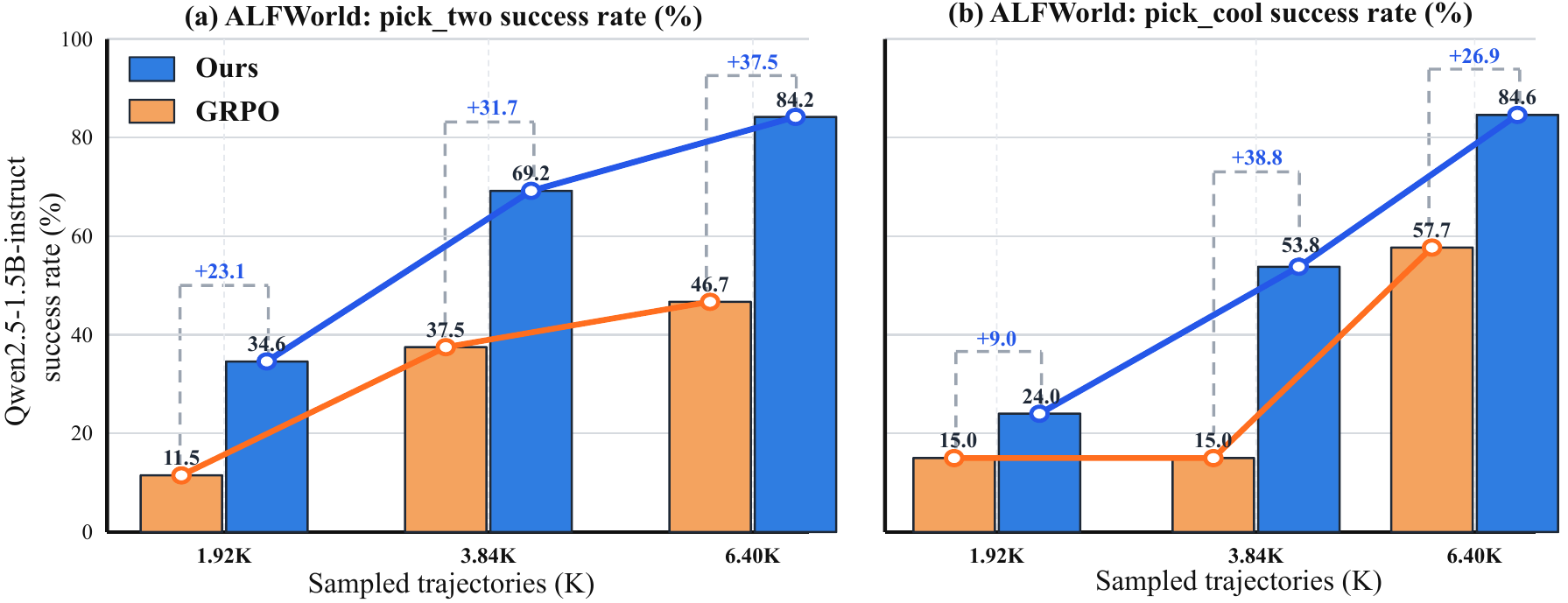}
    \caption{Learning dynamics and sample efficiency of our method with Qwen2.5-1.5B-Instruct. (a) and (b) compare the best success rates achieved by our method and GRPO on the ALFWorld \textit{pick\_two} and \textit{pick\_cool} subtasks.}
    \label{fig:placeholder}
    \vspace{-0.2cm}
\end{figure}



\subsection{Performance on ALFWorld and WebShop}
\label{sec:ALFWorld_webshop_results}
Table~\ref{tab:main_results} shows that TRCA achieves the strongest overall performance at both model scales. With Qwen2.5-1.5B-Instruct, TRCA reaches $92.0\%$ overall of ALFWorld success rate,$90.5\%$ of WebShop task score,  and $78.4\%$ WebShop success rate. These results represent $5.3\%$, $7.4\%$, and $13.4\%$ improvements over GiGPO, respectively. With Qwen2.5-7B-Instruct, TRCA obtains $94.5\%$, $92.9\%$, and $83.8\%$ on the same metrics, obtains $3.7\%$, $8.5\%$, and $11.0\%$ improvements over GiGPO. TRCA also improves upon the strongest prior overall baseline, under Qwen2.5-1.5B-instruct setting, our method achieves $3.8\%$ task score and $2.9\%$ success-rate improvements on WebShop over GraphGPO. under Qwen2.5-7B-instruct setting, the corresponding gains $6.0\%$ and $3.5\%$ improvements over GraphGPO. 


\subsection{Performance on Search-Augmented QA}
As shown in Table~\ref{tab:qa_results}, TRCA achieves the best average performance at both model scales. With Qwen2.5-3B-Instruct, TRCA obtains an average score of $45.4\%$, and it obtains $8.2\%$ and $3.3\%$ improvements over IGPO and GiGPO, respectively. It also achieves the best result on each of the seven individual datasets in the 3B setting. With Qwen2.5-7B-Instruct, TRCA also reaches the best average score over other baselines, which shows that TRCA maintains competitive performance across heterogeneous search tasks and provides the strongest aggregate result without relying on ground truth labels on every dataset.
\begin{figure}[t]
    \centering
    \includegraphics[width=\linewidth]{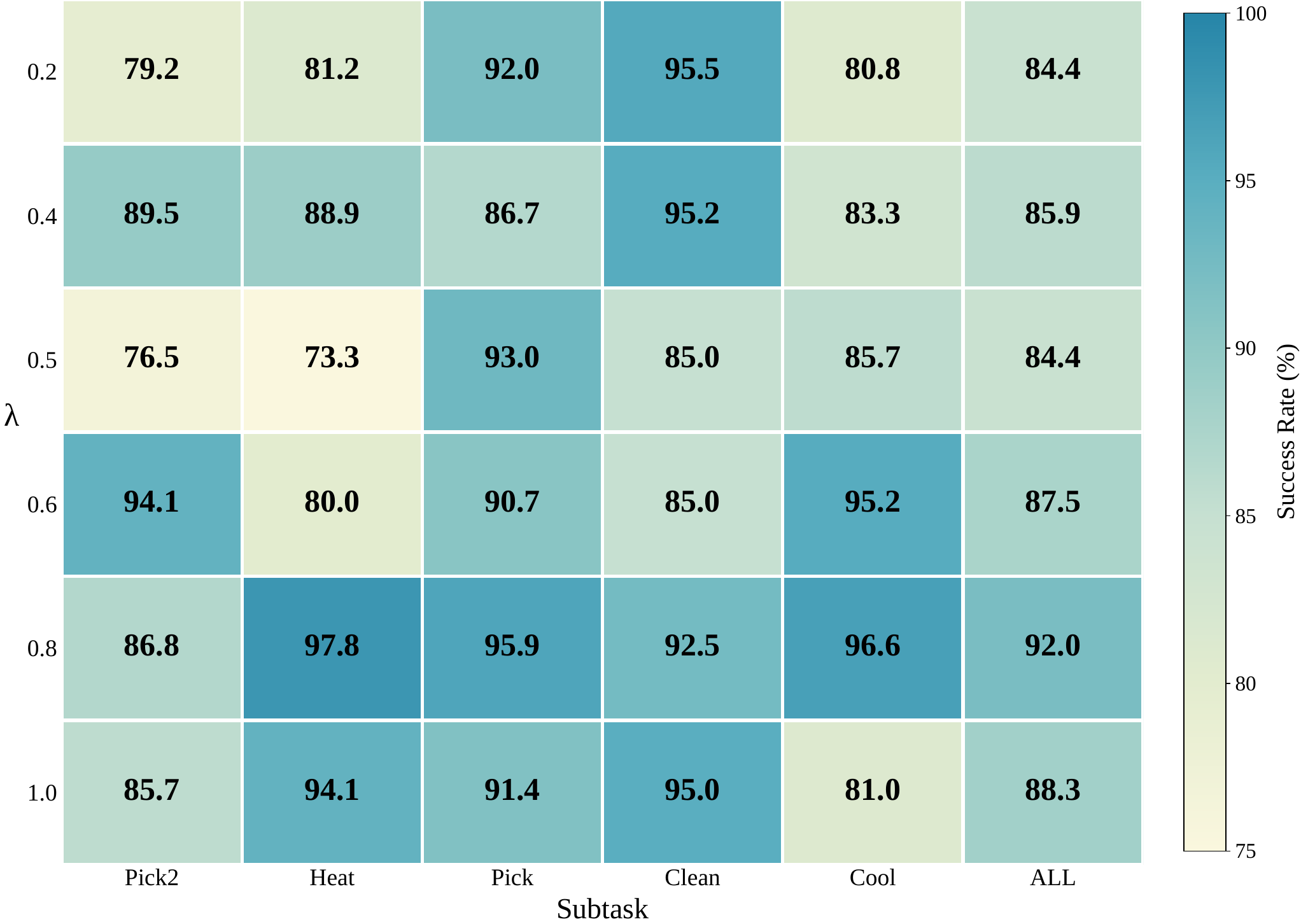}
    \caption{Performance of TRCA under different settings
of hyperparameter $\lambda$ on ALFWorld success rate (\%).}
    
    \label{fig:trca_lambda}
    \vspace{-0.3cm}
\end{figure}

\subsection{Learning Dynamics and Design Analysis}
\label{sec:training_ablation_analysis}
\paragraph{Learning Dynamics and Sample Efficiency.} 
Figure~\ref{fig:placeholder} (a) and (b) show that TRCA converges substantially faster than GRPO on both \textit{Pick\_two} and \textit{Pick\_cool} under identical trajectory-sampling budgets. Notably, this advantage emerges from the earliest stage of training: with only 1.92K sampled trajectories, TRCA improves the success rate from 11.5\% to 34.6\% on \textit{Pick\_two} and from 15.0\% to 24.0\% on \textit{Pick\_cool}. On \textit{Pick\_two}, TRCA reaches a 69.2\% success rate with 3.84K trajectories, surpassing the 46.7\% achieved by GRPO with 6.40K trajectories. On \textit{Pick\_cool}, TRCA achieves 53.8\% with 3.84K trajectories, approaching GRPO's 57.7\% while using 40\% fewer samples. These results demonstrate that transition-wise rubric rewards improve the utilization of early-stage rollout experience, thereby accelerating convergence and enhancing overall sample efficiency. Experimental details are provided in Appendix ~\ref{app:experimental_details}.
\paragraph{Ablation Study.}
We ablate the two reward components of TRCA while keeping all other training configurations unchanged. As shown in Figure~\ref{fig:trca_ablation_1x4_front50_noisy}, removing either component consistently degrades performance, confirming their complementarity. Removing Breakthrough Rubric Reward causes a larger drop across all three benchmarks, indicating that newly covered rubric conditions provide particularly important credit for task-critical actions, while Foundational Rubric Reward supplies broad step-level supervision.

\paragraph{Hyperparameter sensitivity analysis.}
On ALFWorld, we vary \(\lambda\) over \(\{0.2,0.4,0.5,0.6,0.8,1.0\}\), where larger values place
greater emphasis on Breakthrough Rubric Reward. Increasing
\(\lambda\) generally benefits challenging subtasks such as
\textit{Heat} and \textit{Cool}, but excessive emphasis on
breakthrough progress reduces the overall success rate from
\(92.0\%\) at \(\lambda=0.8\) to \(88.3\%\) at \(\lambda=1.0\).
Thus, \(\lambda=0.8\) provides the best empirical balance
between broad step-level supervision and breakthrough-focused
credit.


\vspace{-0.1cm}
\section{Conclusion}
\label{sec:conclusion_limitation}
Our central insight is that failure does not imply the absence of supervision: even unsuccessful trajectories contain abundant and verifiable evidence in the state transitions they induce. TRCA leverages this evidence by grounding credit assignment in environment-observed changes rather than rare terminal successes, thereby converting otherwise discarded experience into reliable step-level supervision without successful anchors or learned process reward models. This capability is particularly valuable during early exploration, when successful rollouts are scarce but informative transitions remain plentiful. Across embodied interaction, web navigation, and search-augmented reasoning, TRCA consistently improves performance, convergence speed, and sample efficiency, especially under limited rollout budgets. More broadly, our results establish transition-level environmental evidence as a general foundation for learning long-horizon behaviors from both successful and failed experience.

\section*{Acknowledgments}

This work was supported in part by the Beijing Major Science and Technology
Project under Contract No.~Z251100008125031 and the National Natural Science
Foundation of China under Grant No.~62401327. This work was also supported
by the Beijing Academy of Artificial Intelligence (BAAI).

\bibliography{main}
\newpage
\twocolumn

\clearpage
\appendix
\newpage
\setcounter{equation}{0}
\setcounter{section}{0}
\setcounter{secnumdepth}{2}
\renewcommand{\thesection}{\Alph{section}}
\renewcommand{\thesubsection}{\thesection.\arabic{subsection}}

\section{Task-Conditioned Rubric Construction and Evaluation}
\label{app:rubric_construction}

\paragraph{Overview of rubric construction and evaluation.}
Appendix~A complements the rubric formulation in the main
text by specifying how a shared rubric library is constructed,
instantiated for a concrete task, and applied to an
action-induced transition. Before policy training, we use GPT-5.6-sol to construct a benchmark-specific rubric operator library from a shared three-dimensional rubric schema comprising Evidence, Invalidity, and Execution. For each benchmark, the model is invoked once with benchmark-specific interface information to generate 15 deterministic Boolean operators, with five operators per category. The resulting library remains fixed throughout policy training and is shared across all task instances within the same benchmark. Task-specific requirements are subsequently incorporated through deterministic parameter binding, while transition-level rubric judgments are computed without additional LLM inference. At the beginning, we let
\[
b\in
\{
\textsc{ALFWorld},
\textsc{WebShop},
\textsc{SearchQA}
\}
\]
index the benchmark-specific interaction type, and let
\(x\in\mathcal{X}_b\) denote a task instance from benchmark
\(b\).

For rubric category
\(c\in\{\mathrm{Evi},\mathrm{Inval},\mathrm{Exec}\}\), TRCA
uses a benchmark-level operator \(h_{b,m}^{c}\), a
task-conditioned binding \(\omega_{b,m}(x)\), and a
benchmark-specific transition adapter \(g_b\). Their
composition produces the binary judgment
\begin{equation}
\mathcal{O}_{i,t,(m,\omega)}
=
h_{b,m}^{c}
\left(
\omega_{b,m}(x),
g_b(s_{i,t},a_{i,t},s_{i,t+1})
\right)
\in\{0,1\}.
\label{eq:preliminary_rubric_judgment}
\end{equation}

Here, \(h_{b,m}^{c}\) defines a reusable Boolean condition,
\(\omega_{b,m}(x)\) supplies the task-specific parameters
required by that condition, and \(g_b\) maps the raw
action-induced transition into canonical observable facts
available at the current interaction step. The operator is
active only when these observable facts satisfy its Boolean
condition; missing, ambiguous, or unsupported evidence leaves
the operator inactive.

\subsection{Shared Rubric Schema and Benchmark-Level
Operator Libraries}
\label{app:benchmark_rubric_instantiation}

At the level of rubric design, ALFWorld, WebShop, and
SearchQA are treated as instances of a common long-horizon
interaction family. They share the same three-category rubric
schema and the same library-generation protocol:
\textsc{Evidence}, \textsc{Invalidity}, and
\textsc{Execution}. The schema fixes the semantic role of
each category across benchmarks, while the concrete operators
are instantiated according to the action grammar and
observable feedback exposed by benchmark \(b\).

Before policy training, an LLM is invoked only once for each
benchmark-specific interaction type \(b\) to construct the
corresponding reusable operator library.
\begin{equation}
\mathcal{T}_{b}
=
\left\{
\mathcal{T}_{b}^{\mathrm{Evi}},
\mathcal{T}_{b}^{\mathrm{Inval}},
\mathcal{T}_{b}^{\mathrm{Exec}}
\right\},
\qquad
\mathcal{T}_{b}^{c}
=
\left\{
h_{b,m}^{c}
\right\}_{m=1}^{M_c},
\qquad
\label{eq:app_benchmark_operator_library}
\end{equation}
where
\(c\in\{\mathrm{Evi},\mathrm{Inval},\mathrm{Exec}\}\).
Each library therefore contains five operators per category
and fifteen operators in total.

The three categories have complementary roles.
\textsc{Evidence} operators identify newly revealed or newly
supported task-relevant entities, attributes, locations,
relations, or supporting facts. \textsc{Invalidity}
operators identify malformed, unavailable, rejected,
repeated, no-op, or precondition-violating actions.
\textsc{Execution} operators clearly identify accepted
task-relevant operations, successfully achieved subgoals,
completed state transformations, or explicitly satisfied
task conditions.

Each operator \(h_{b,m}^{c}\) specifies: (i) the canonical
observable facts it reads, (ii) the task-binding fields it
requires, and (iii) an exact Boolean activation condition.
An operator returns one only when its required observable
facts establish the condition; Otherwise, it returns zero. This fully deterministic
form effectively prevents unsupported or ambiguous transition
evidence from producing positive rubric judgments.

The same generation prompt, category definitions, output
schema, and Boolean-writing requirements are used for all
three benchmarks. Benchmark-specific information is limited
to the interface description required to define executable
operators, including the action grammar, admissible
operations, observable state fields, and environment feedback
format. Consequently, the generated libraries adapt to
different interaction interfaces without changing the
high-level rubric design. Library construction is performed once before training and
does not use successful trajectories, terminal outcomes,
future transitions, task-specific answers, benchmark
performance results, or ground-truth intermediate labels.
The LLM constructs only the reusable benchmark-level
operators. It is not invoked to generate a new library for an
individual task and is not used as a transition judge during
rollout collection or policy optimization.
Once \(\mathcal{T}_{b}\) has been constructed, no further LLM
calls are required. For each task, the deterministic binding
procedure in Appendix~\ref{app:shared_rubrics} fills the
operator fields and produces the grounded rubric set
\(\mathcal{R}(x)\). The resulting task-grounded operators are
evaluated on action-induced transitions as described in
Appendix~\ref{app:rule_based_evaluation}, using the
environment-specific observable signals summarized in
Appendix~\ref{app:environment_signals}. Their binary outputs
are converted into Foundational and Breakthrough Rubric
Rewards in Appendix~\ref{app:reward_construction}. The shared
generation prompt and benchmark-level libraries are provided
in Appendix~\ref{app:rubric_prompt}.

\subsection{Task-Conditioned Binding across Task Instances}
\label{app:shared_rubrics}

All task instances within benchmark-specific interaction type
\(b\) reuse the same operator library \(\mathcal{T}_{b}\).
Each operator \(h_{b,m}^{c}\) is a parameterized Boolean
template with predefined binding fields. For a concrete task
instruction \(x\in\mathcal{X}_{b}\), a deterministic task
binder fills these fields with task-specific parameters:
\begin{equation}
\omega_{b,m}(x)
=
\mathcal{B}_{b,m}(x).
\label{eq:app_task_binding}
\end{equation}
The binder is implemented through benchmark-specific parsing
and field extraction and does not invoke an LLM or another
learned model. This stage performs deterministic parameter
binding rather than generating a new rubric library.

The binding \(\omega_{b,m}(x)\) may contain task-relevant
entities, attributes, relations, locations, constraints,
required operation types, target conditions, or output
formats extracted from the task instruction and fixed
benchmark metadata. For operators that depend only on
benchmark-level interface properties, such as parser
validity, the corresponding task binding may be empty. The grounded rubric set for task \(x\) is defined as
\begin{equation}
\mathcal{R}^{c}(x)
=
\left\{
\left(
h_{b,m}^{c},
\omega_{b,m}(x)
\right)
\right\}_{m=1}^{M_c},
\qquad
x\in\mathcal{X}_{b},
\label{eq:app_grounded_rubrics}
\end{equation}
where
\(c\in\{\mathrm{Evi},\mathrm{Inval},\mathrm{Exec}\}\).
The operator semantics \(h_{b,m}^{c}\) remain fixed across
all task instances of benchmark \(b\); only the
task-conditioned bindings \(\omega_{b,m}(x)\) vary with the
specific instruction. Consequently, no new operator library
is separately generated for any particular task instance.

Accordingly, the task-conditioned rubric set
\(\mathcal{R}^{c}(x)\) consists of concrete rubric items
instantiated from the shared operators through deterministic
parameter binding; it is not an independently generated
task-specific library.

For example, the ALFWorld binder extracts the target object,
required state transformation, quantity, and destination
receptacle. The WebShop binder extracts requested product
attributes, options, and purchase constraints. The SearchQA
binder extracts the target entity or subject, requested
relation or information need, query constraints, and required
answer format. Ground-truth answer aliases are not included
in the SearchQA binding and are used only by the environment
to compute the terminal Exact Match outcome.

The task binding remains fixed within a task instance, while
the observable facts supplied to the bound operators vary
across action-induced transitions.

\subsection{Rule-Based Transition Evaluation}
\label{app:rule_based_evaluation}

For rollout \(i\), the transition at step \(t\) is
represented as
\begin{equation}
\xi_{i,t}
=
\left(
s_{i,t},
a_{i,t},
s_{i,t+1}
\right),
\label{eq:app_transition}
\end{equation}
where \(s_{i,t}\) is the current interaction state,
\(a_{i,t}\) is the complete environment-facing action, and
\(s_{i,t+1}\) is the next interaction state containing the
observable environment feedback produced by executing the
action.

During training, the deterministic benchmark adapter \(g_b\)
maps the raw action-induced transition into canonical
observable facts. For a task-grounded rubric item
\[
j
=
\left(
h_{b,m}^{c},
\omega_{b,m}(x)
\right)
\in\mathcal{R}^{c}(x),
\]
the transition-level judgment is computed as
\begin{equation}
\mathcal{O}_{i,t,j}
=
h_{b,m}^{c}
\left(
\omega_{b,m}(x),
g_b(s_{i,t},a_{i,t},s_{i,t+1})
\right)
\in\{0,1\}.
\label{eq:app_rubric_judgment}
\end{equation}

The binding \(\omega_{b,m}(x)\) supplies the task-specific
requirements, while \(g_b\) supplies the observable facts
associated with the current pre-action state, action, and
post-action environment feedback. These facts may include
the parsed action type and arguments, parser status,
currently available operations, visible entities and
attributes, retrieved content, action acceptance or rejection
feedback, and observable state changes caused by the action.

The evaluation does not access successful trajectories,
terminal labels, future states, hidden environment states,
benchmark performance results, or ground-truth intermediate
labels. If the observable evidence is insufficient to
establish a rubric condition, the corresponding operator
remains inactive and
\(\mathcal{O}_{i,t,j}=0\). Here, zero denotes non-activation,
rather than negative reward. Negative credit is introduced
only when an \textsc{Invalidity} operator is activated and
mapped to its signed contribution by Foundational Rubric
Reward.

\textsc{Evidence} judgments are triggered by newly revealed
or newly supported task-relevant information, such as an
object location in ALFWorld, a product attribute in WebShop,
or a supporting fact in SearchQA. \textsc{Invalidity}
judgments are triggered by parser failure, malformed command
format, environment rejection, inadmissible action,
unavailable interface operation, repeated ineffective
execution, or another directly observable failure.
\textsc{Execution} judgments are triggered when the
transition satisfies a task-required operation, subgoal, or
intermediate condition, such as acquiring an object,
selecting an instruction-consistent option, or issuing an
accepted retrieval step within the current interaction
context.

A single transition can satisfy multiple rubric items within
the same category or across different categories. For
example, an action may both reveal a task-relevant object and
complete a required exploration operation. The resulting
Boolean judgments are evaluated independently and are
subsequently aggregated by the reward construction in
Appendix~\ref{app:reward_construction}.

Both task binding and transition evaluation are deterministic.
Neither stage invokes an LLM during rollout collection or
policy optimization. Accordingly, the LLM is used exclusively for the one-time
benchmark-level operator library construction process described in Appendix~\ref{app:benchmark_rubric_instantiation}.

\subsection{Environment-Specific Observable Signals}
\label{app:environment_signals}

The following signals are the transition-dependent observable
facts extracted by the benchmark adapter \(g_b\). They are evaluated together with the corresponding
task-conditioned bindings formally defined in Appendix~\ref{app:shared_rubrics}.

\paragraph{ALFWorld.}
ALFWorld exposes textual feedback concerning action
admissibility, object locations, inventory states, receptacles,
and object-state transformations. \textsc{Evidence} items
detect newly observed task-relevant objects, locations,
receptacles, or object states. \textsc{Invalidity} items
detect malformed commands, inadmissible actions, unavailable
objects, unsatisfied preconditions, environment rejection,
repeated ineffective actions, or failed execution.
\textsc{Execution} items identify task-relevant operations
such as acquiring an object, applying a required
transformation, or placing an object in a target receptacle
throughout the current interaction process.

\paragraph{WebShop.}
WebShop provides structured page states, product attributes,
available options, selected options, navigation feedback, cart
state, and purchase status. \textsc{Evidence} items detect
newly revealed products, attributes, or task-relevant options.
\textsc{Invalidity} items identify malformed interface
actions, unavailable selections, rejected operations,
repeated no-op actions, or premature purchases.
\textsc{Execution} items identify substantive operations such
as opening a candidate product, selecting an
instruction-consistent option, adding a qualifying product to
the cart, or reaching a purchase-ready state during the
current  navigation process.

\paragraph{SearchQA.}
SearchQA exposes search queries, retrieved results, document
content, tool-execution feedback, retrieval-state changes, and
answer-submission states. \textsc{Evidence} items detect newly
retrieved task-relevant entities, relations, passages, or
supporting facts. \textsc{Invalidity} items detect malformed
tool calls, unavailable result or passage identifiers,
unsatisfied action preconditions, rejected queries, repeated
no-op retrieval actions, or invalid interaction formats.
\textsc{Execution} items identify accepted operations such as
issuing a valid search, opening a retrieved source, reading a
relevant passage, extracting supporting evidence, or
submitting an answer in the required format during the
current process.

SearchQA operators may inspect retrieved text to determine
whether it expresses a relation relevant to the task
instruction. They do not compare the retrieved text or the
submitted answer against a ground-truth answer alias.
Ground-truth answer aliases are used only by the benchmark to
compute the terminal Exact Match reward and are not used to
judge intermediate rubric items.

\subsection{Validity in the Diagnostic Study}
\label{app:validity_diagnostic}

Figure~\ref{fig:Motivation} reports \textsc{Validity} as a
diagnostically useful transition signal because successful
parsing and execution are directly verifiable from environment
feedback. Although a valid action may not immediately satisfy
a task subgoal, it provides reliable evidence that the agent
can interact with the environment in an executable way.

In the TRCA reward construction, we model the complementary
failure condition, \textsc{Invalidity}, and assign negative
credit to malformed, inadmissible, rejected, repeated, or
otherwise unexecutable actions. Thus, \textsc{Validity} in the
diagnostic analysis and \textsc{Invalidity} in the reward
construction are complementary views of action executability.

The absence of an Invalidity judgment does not itself produce
positive reward. A valid and executable action receives
positive rubric credit only when the same transition also
activates an \textsc{Evidence} or \textsc{Execution} item.
Thus, positive supervision remains tied to task-relevant
information acquisition or substantive task execution rather
than action validity alone. Together, Appendices~A.1--A.5 specify the complete rubric
instantiation and evaluation pipeline. Appendix~A.1 constructs
the shared benchmark-level operator library
\(\mathcal{T}_{b}\). Appendix~A.2 deterministically binds its
parameterized operators to a task instruction, producing the
concrete rubric set \(\mathcal{R}(x)\). Appendix~A.3 applies
these task-grounded operators to observable action-induced
transitions to obtain the binary judgments
\(\mathcal{O}_{i,t,j}\), while Appendices~A.4 and~A.5 specify
their observable inputs and executability semantics. These
judgments are subsequently converted into Foundational and
Breakthrough Rubric Rewards in
Appendix~\ref{app:reward_construction}.

\section{Notation and TRCA Training Algorithm}
\label{app:notation_algorithm}

\subsection{Notation Summary}
\label{app:notation}

For a task-grounded rubric item
\(j=(h_{b,m}^{c},\omega_{b,m}(x))\), its binary judgment is
computed as
\begin{equation}
\mathcal{O}_{i,t,j}
=
h_{b,m}^{c}\left(
\omega_{b,m}(x),
g_b\left(s_{i,t},a_{i,t},s_{i,t+1}\right)
\right)
\in\{0,1\},
\label{eq:app_preliminary_rubric_judgment}
\end{equation}
where \(h_{b,m}^{c}\) is a shared operator in category
\(c\in\{\mathrm{Evi},\mathrm{Inval},\mathrm{Exec}\}\),
\(\omega_{b,m}(x)\) is its deterministic task-conditioned
binding, and \(g_b\) is the benchmark-specific transition
adapter. Table~\ref{tab:notation} summarizes the main notation used
in the rubric construction and transition-wise credit
assignment of TRCA throughout the complete training pipeline;
standard reinforcement-learning notation is omitted for
brevity.

\begin{table}[t]
\centering
\footnotesize
\caption{Main notation used in TRCA.}
\label{tab:notation}
\renewcommand{\arraystretch}{1.08}
\setlength{\tabcolsep}{3pt}
\begin{tabular}{p{0.17\columnwidth} p{0.75\columnwidth}}
\toprule
\textbf{Symbol} & \textbf{Definition} \\
\midrule

\(x\) &
A concrete task instruction or query. \\

\(b\) &
A benchmark-specific interaction type, where
\(b\in\{\textsc{ALFWorld},\textsc{WebShop},
\textsc{SearchQA}\}\). \\

\(\mathcal{T}_{b}\) &
The benchmark-level operator library shared across all task
instances of benchmark \(b\), consisting of Evidence,
Invalidity, and Execution operators. \\

\(h_{b,m}^{c}\) &
The \(m\)-th deterministic Boolean operator in category \(c\)
of library \(\mathcal{T}_{b}\). \\

\(\xi_{i,t}\) &
The action-induced transition
\(\xi_{i,t}=(s_{i,t},a_{i,t},s_{i,t+1})\)
at interaction step \(t\) of rollout \(i\). \\

\(\omega_{b,m}(x)\) &
The task-conditioned parameters deterministically bound to
operator \(h_{b,m}^{c}\), such as an entity, attribute,
relation, constraint, target condition, or required output
format. \\

\(g_b\) &
The benchmark-specific adapter that maps the raw transition
\((s_{i,t},a_{i,t},s_{i,t+1})\)
into canonical observable facts. \\

\(\mathcal{R}^{c}(x)\) &
The grounded rubric-item set formed by pairing the shared
operators in \(\mathcal{T}_{b}^{c}\) with their
task-conditioned bindings for task \(x\). \\

\(\mathcal{O}_{i,t,j}\) &
The binary judgment
\(\mathcal{O}_{i,t,j}\in\{0,1\}\)
indicating whether transition \(\xi_{i,t}\) satisfies rubric item \(j\). \\

\(r^{\mathrm{F}}_{i,t}\) &
The Foundational Rubric Reward obtained by aggregating the signed rubric judgments at transition \((i,t)\). \\

\(C_{i,t,j}\) &
The cumulative coverage indicator specifying whether positive rubric item \(j\) has been satisfied by any transition up to step \(t\). \\

\(r^{\mathrm{B}}_{i,t}\) &
The Breakthrough Rubric Reward assigned to newly covered Evidence and Execution items. \\

\(r^{\mathrm{TRCA}}_{i,t}\) &
The transition-level reward obtained by combining Foundational Rubric Reward and Breakthrough Rubric Reward. \\

\(\lambda\) &
The mixing coefficient between Foundational Rubric Reward and Breakthrough Rubric Reward. \\

\bottomrule
\end{tabular}
\end{table}

\subsection{Normalization and Numerical Conventions}
\label{app:normalization_conventions}

TRCA uses normalization only when constructing relative
advantages. Reward normalization and advantage normalization
are therefore separate operations. The rubric rewards
\(r^{\mathrm{F}}_{i,t}\), \(r^{\mathrm{B}}_{i,t}\), and
\(r^{\mathrm{TRCA}}_{i,t}\) are defined by the category
budgets and the item-count normalization in
Appendix~\ref{app:reward_construction}. The episode-relative
and context-relative advantages are normalized over the
corresponding rollout group or context group.

For a finite comparison group \(\mathcal{G}\) of scalar values
\(\{z_k\}\), we consistently use the following normalization
convention
\begin{equation} 
\operatorname{Norm}_{\mathcal G}(z_k)
=
\begin{cases}
\dfrac{z_k-\mu_{\mathcal G}}
{\sigma_{\mathcal G}+\epsilon_{\mathrm{norm}}},
&
\sigma_{\mathcal G}>0,
\\[6pt]
0,
&
\sigma_{\mathcal G}=0.
\end{cases}
\label{eq:app_normalization}
\end{equation}
Here, \(\mu_{\mathcal G}\) and \(\sigma_{\mathcal G}\) are
computed exclusively within the corresponding comparison
group for the current normalization. 

For episode-relative normalization, the group is the rollout
group \(\mathcal{G}(x)=\{\tau_i\}_{i=1}^{N}\), and the
normalized values are terminal outcomes \(R(\tau_i)\). If all
rollouts in the group have the same outcome, such as an
all-failure group or an all-success group, then the standard
deviation is zero and \(A^{\mathrm{E}}(\tau_i)=0\). For
context-relative normalization, the group is
\(\mathcal{G}^{\mathrm{S}}(\widetilde{s})\), and the
normalized values are completion-aware returns \(R_{i,t}\)
from comparable decision contexts. If the context group has
fewer than two usable transitions or has zero variance, then
\(A^{\mathrm{S}}(a_{i,t})=0\). These conventions avoid
introducing artificial preference when the sampled group
contains no distinguishable outcome or return signal.

\subsection{Action and Token Conventions}
\label{app:action_token_conventions}

An interaction step is an environment-facing decision point,
whereas a token is an autoregressive generation unit. The
variable \(a_{i,t}\) denotes the complete agent action emitted
at interaction step \(t\), not a single token. In the
experiments, the model generates reasoning in \texttt{<think>}
tags and the executable command in \texttt{<action>} tags. The
environment transition \(\xi_{i,t}\) is induced by the
executable action and the subsequent observable environment
feedback.

For each interaction step, TRCA computes one scalar transition
advantage for the complete environment action. During
token-level policy optimization, this advantage is broadcast
only to the generated executable tokens inside the
\texttt{<action>}\(\ldots\)\texttt{</action>} span.
Reasoning tokens inside
\texttt{<think>}\(\ldots\)\texttt{</think>}, prompt tokens,
observation tokens, retrieved-content tokens, padding tokens,
and tokens belonging to other interaction steps are masked out.

\section{Detailed TRCA Reward Construction}
\label{app:reward_construction}

This section provides additional details for the two
transition-level reward components introduced in the main
paper. The construction follows the same notation and formulas.

\subsection{Foundational Rubric Reward}
\label{app:foundational_reward}

For rubric item \(j\) at transition \((i,t)\), TRCA defines
the signed item-level contribution
\begin{equation}
q^i_{t,j}
=
\begin{cases}
\displaystyle
+\frac{r^{\mathrm{Evi}}}{M_{\mathrm{Evi}}},
&
j\in\mathcal{R}^{\mathrm{Evi}}(x),
\quad
\mathcal{O}_{i,t,j}=1,
\\[6pt]
\displaystyle
-\frac{\lvert r^{\mathrm{Inval}}\rvert}
{M_{\mathrm{Inval}}},
&
j\in\mathcal{R}^{\mathrm{Inval}}(x),
\quad
\mathcal{O}_{i,t,j}=1,
\\[6pt]
\displaystyle
+\frac{r^{\mathrm{Exec}}}{M_{\mathrm{Exec}}},
&
j\in\mathcal{R}^{\mathrm{Exec}}(x),
\quad
\mathcal{O}_{i,t,j}=1,
\\[6pt]
0,
&
\text{otherwise},
\end{cases}
\label{eq:app_item_contribution}
\end{equation}
where \(M_c=|\mathcal{R}^{c}(x)|\) is the number of rubric
items in category \(c\). Evidence and Execution contribute
positive credit, while Invalidity contributes negative credit.
If no rubric item is supported by observable transition
evidence, then all corresponding item contributions are zero.

The Foundational Rubric Reward aggregates all item-level
contributions at the same transition:
\begin{equation}
r^{\mathrm{F}}_{i,t}
=
\sum_{j\in\mathcal{R}(x)}
q^i_{t,j}.
\label{eq:app_foundational_reward}
\end{equation}
A transition may satisfy more than one rubric item, so
Equation~\eqref{eq:app_foundational_reward} sums over all
triggered Evidence, Invalidity, and Execution judgments.
Dividing each category budget by \(M_c\) bounds the magnitude
of that category's aggregate contribution at any transition by
its predefined reward budget. Consequently, the reward scale
does not grow simply because a benchmark uses a longer rubric
list.

Foundational Rubric Reward evaluates the current transition
locally. The same rubric item can contribute at multiple
steps if distinct transitions repeatedly provide observable
support for that item. This local behavior is intentional:
Foundational Rubric Reward supplies broad transition-level
supervision over informative, invalid, and task-relevant
actions, while Breakthrough Rubric Reward, defined next,
separately tracks newly covered positive conditions.

\subsection{Breakthrough Rubric Reward}
\label{app:breakthrough_reward}

To distinguish newly achieved progress from repeatedly
observed evidence, TRCA defines the cumulative coverage of
item \(j\) as
\begin{equation}
C_{i,t,j}
=
\max_{1\le k\le t}
\mathcal{O}_{i,k,j},
\qquad
C_{i,0,j}=0.
\label{eq:app_rubric_coverage}
\end{equation}
The cumulative potential over positive rubric items is
\begin{equation}
\Phi_{i,t}
=
\sum_{c\in\{\mathrm{Evi},\mathrm{Exec}\}}
\frac{r^{c}}{M_c}
\sum_{j\in\mathcal{R}^{c}(x)}
C_{i,t,j}.
\label{eq:app_rubric_potential}
\end{equation}
Invalidity items are excluded from this positive potential
because they describe transition-local errors and are already
penalized by Foundational Rubric Reward. Under the coverage
definition in Equation~\eqref{eq:app_rubric_coverage},
\(\Phi_{i,t}\) is monotone nondecreasing in \(t\), assuming
nonnegative Evidence and Execution budgets.

The Breakthrough Rubric Reward is the temporal difference of
this cumulative potential:
\begin{equation}
r^{\mathrm{B}}_{i,t}
=
\Phi_{i,t}
-
\Phi_{i,t-1}.
\label{eq:app_breakthrough_difference}
\end{equation}
Equivalently,
\begin{equation}
r^{\mathrm{B}}_{i,t}
=
\sum_{c\in\{\mathrm{Evi},\mathrm{Exec}\}}
\frac{r^{c}}{M_c}
\sum_{j\in\mathcal{R}^{c}(x)}
\left(
1-C_{i,t-1,j}
\right)
\mathcal{O}_{i,t,j}.
\label{eq:app_breakthrough_expanded}
\end{equation}
Thus, rubric item \(j\) contributes to
\(r^{\mathrm{B}}_{i,t}\) only when the current transition
satisfies that item and no earlier transition in the same
rollout has covered it. Already covered items do not receive
repeated Breakthrough Rubric Reward. Since the potential is
cumulative and uses a maximum over previous judgments, the
current BRR definition does not introduce a coverage-regression
penalty. When Evidence and Execution budgets are nonnegative,
\(r^{\mathrm{B}}_{i,t}\ge 0\), and the cumulative BRR over a
rollout is bounded by \(r^{\mathrm{Evi}}+r^{\mathrm{Exec}}\).

\subsection{Reward Fusion}
\label{app:reward_fusion}

The two reward components are combined as
\begin{equation}
r^{\mathrm{TRCA}}_{i,t}
=
(1-\lambda)
r^{\mathrm{F}}_{i,t}
+
\lambda
r^{\mathrm{B}}_{i,t},
\qquad
\lambda\in[0,1].
\label{eq:app_trca_reward}
\end{equation}
Here, \(\lambda\) controls the trade-off between local broad
supervision and newly covered positive rubric progress. When
\(\lambda=0\), TRCA uses only Foundational Rubric Reward. When
\(\lambda=1\), it uses only Breakthrough Rubric Reward. We set \(\lambda=0.8\) in all experiments unless otherwise
specified.

Foundational Rubric Reward provides both positive and negative
local supervision, including penalties for Invalidity.
Breakthrough Rubric Reward highlights transitions that
introduce previously uncovered Evidence or Execution
conditions. Their convex combination preserves the method's
two intended sources of transition-wise credit without
requiring successful anchors or learned process evaluators.

\begin{table*}[t]
\centering
\small
\caption{Symbolic illustration of TRCA reward construction.
The example uses symbolic budgets and does not report an
experimental trajectory.}
\label{tab:app_worked_example}
\begin{tabular}{p{0.11\linewidth}p{0.24\linewidth}p{0.24\linewidth}p{0.31\linewidth}}
\toprule
Step & Transition evidence & Triggered judgments & Reward computation \\
\midrule
1 &
The agent inspects a receptacle and observes the apple. &
\(\mathcal{O}_{1,e_{\mathrm{loc}}}=1\). &
\(r^{\mathrm{F}}_{1}=r^{\mathrm{Evi}}/M_{\mathrm{Evi}}\).
Coverage changes from \(0\) to \(1\) for \(e_{\mathrm{loc}}\), so
\(r^{\mathrm{B}}_{1}=r^{\mathrm{Evi}}/M_{\mathrm{Evi}}\). \\
2 &
The agent takes the apple. &
\(\mathcal{O}_{2,u_{\mathrm{take}}}=1\). &
\(r^{\mathrm{F}}_{2}=r^{\mathrm{Exec}}/M_{\mathrm{Exec}}\).
The item \(u_{\mathrm{take}}\) is newly covered, so
\(r^{\mathrm{B}}_{2}=r^{\mathrm{Exec}}/M_{\mathrm{Exec}}\). \\
3 &
The agent repeats an observation that supports
\(e_{\mathrm{loc}}\). &
\(\mathcal{O}_{3,e_{\mathrm{loc}}}=1\). &
\(r^{\mathrm{F}}_{3}=r^{\mathrm{Evi}}/M_{\mathrm{Evi}}\).
Since \(C_{2,e_{\mathrm{loc}}}=1\), the repeated item gives
\(r^{\mathrm{B}}_{3}=0\). \\
4 &
The environment rejects a malformed command. &
\(\mathcal{O}_{4,v_{\mathrm{bad}}}=1\). &
\(r^{\mathrm{F}}_{4}=-|r^{\mathrm{Inval}}|/M_{\mathrm{Inval}}\).
Invalidity is excluded from \(\Phi\), so \(r^{\mathrm{B}}_{4}=0\). \\
\bottomrule
\end{tabular}
\end{table*}

\section{Training Algorithm and Policy Optimization}
\label{app:training_algorithm}

\subsection{Completion-Aware Return}
\label{app:completion_return}

The environment reward is typically sparse in the paper's
long-horizon settings: \(r_{i,t}=0\) for nonterminal steps and
the terminal transition receives the task outcome
\(R(\tau_i)\). TRCA augments this sparse environment reward
with the transition-level rubric signal for constructing the final reward:
\begin{equation}
R_{i,t}
=
\sum_{k=t}^{T_i}
\gamma^{k-t}
\left(
r_{i,k}
+
r^{\mathrm{TRCA}}_{i,k}
\right).
\label{eq:app_completion_return}
\end{equation}
The return starts at the current interaction step and
accumulates future terminal and transition-level signals. We use a discount factor of \(\gamma=0.95\) in all experiments.

The terminal outcome is included through the original
environment reward \(r_{i,k}\), while TRCA adds separate
transition-level supervision through \(r^{\mathrm{TRCA}}_{i,k}\).
The construction therefore does not duplicate the terminal
reward; it combines the sparse completion signal with
observable action-induced transition evidence before relative
advantage normalization.

\subsection{Zero-Variance Rollout Groups}
\label{app:zero_variance_episode}

For a rollout group
\(\mathcal{G}(x)=\{\tau_i\}_{i=1}^{N}\), the
episode-relative advantage is computed from terminal outcomes:
\begin{equation}
A^{\mathrm{E}}(\tau_i)
=
\begin{cases}
\dfrac{R(\tau_i)-\mu_{\mathcal{G}}}
{\sigma_{\mathcal{G}}+\epsilon_{\mathrm{norm}}},
&
\sigma_{\mathcal{G}}>0,
\\[6pt]
0,
&
\sigma_{\mathcal{G}}=0,
\end{cases}
\label{eq:app_episode_advantage}
\end{equation}
where \(\mu_{\mathcal{G}}\) and \(\sigma_{\mathcal{G}}\)
denote the mean and standard deviation of terminal outcomes
within the rollout group.

Zero variance occurs when sampled rollouts in the group have
indistinguishable terminal outcomes. This includes all-failure
groups, where every rollout has \(R(\tau_i)=0\), and all-success
groups, where every rollout has \(R(\tau_i)=1\). In either
case, terminal outcomes provide no within-group preference, so
TRCA sets \(A^{\mathrm{E}}(\tau_i)=0\) instead of amplifying
numerical noise. The step-level transition supervision then
becomes the primary differentiating signal. This convention
does not imply that every failed rollout is identical in its
intermediate behavior; it only states that terminal outcomes
alone cannot rank rollouts inside a zero-variance group.

\subsection{Context-Based Step Grouping}
\label{app:step_grouping}

Following the standard group-in-group formulation, transitions
arising from comparable decision contexts are consistently
assigned to the same step-level group:
\begin{equation}
\mathcal{G}^{\mathrm{S}}(\widetilde{s})
=
\left\{
(i,t)
\mid
s_{i,t}\simeq\widetilde{s},
\ 1\le i\le N,
\ 1\le t\le T_i
\right\}.
\label{eq:app_step_group}
\end{equation}
The relation \(s_{i,t}\simeq\widetilde{s}\) indicates that the
decision context is comparable under environment-specific
state information. It is used exclusively for relative normalization of
completion-aware returns, rather than as a successful anchor
or any additional independent reward source.

The grouping need not be exact textual matching. It can use
structural fields exposed by the environment, such as page
type, product identity, selected options, task predicates,
object state, inventory, tool type, query state, or retrieval
stage. Appendix~\ref{app:context_grouping_details} lists the
benchmark-specific fields used at this level of description.
Context grouping uses only observable benchmark-specific
state fields and does not require embedding models, learned
state encoders, or additional online LLM inference. The context-relative step advantage is
\begin{equation}
A^{\mathrm{S}}(a_{i,t})
=
\begin{cases}
\dfrac{
R_{i,t}
-
\mu_{\mathcal{G}^{\mathrm{S}}}
}{
\sigma_{\mathcal{G}^{\mathrm{S}}}
+
\epsilon_{\mathrm{norm}}
},
&
\sigma_{\mathcal{G}^{\mathrm{S}}}>0,
\\[6pt]
0,
&
\sigma_{\mathcal{G}^{\mathrm{S}}}=0,
\end{cases}
\label{eq:app_step_advantage}
\end{equation}
where the mean and standard deviation are computed over the
completion-aware returns in the corresponding context group
used for the current comparison. If a group is too small to
support comparison or has zero variance, TRCA simply sets the
corresponding step advantage to zero. The final combined
advantage is
\begin{equation}
A^{\mathrm{TRCA}}_{i,t}
=
A^{\mathrm{E}}(\tau_i)
+
A^{\mathrm{S}}(a_{i,t}).
\label{eq:app_final_advantage}
\end{equation}

\subsection{Action-Level Policy Optimization}
\label{app:token_broadcasting}

TRCA treats the complete environment-facing action as the
credit-assignment and policy-optimization unit. For rollout
\(i\) at interaction step \(t\), let \(a_{i,t}\) denote the
complete executable action generated under state \(s_{i,t}\)
and task instruction \(x\). TRCA assigns one scalar transition
advantage \(A_{i,t}^{\mathrm{TRCA}}\) to this action. The corresponding action-level importance ratio as follows.
\begin{equation}
\rho_{i,t}(\theta)
=
\frac{
\pi_{\theta}
\left(
a_{i,t}
\mid
s_{i,t},x
\right)
}{
\pi_{\theta_{\mathrm{old}}}
\left(
a_{i,t}
\mid
s_{i,t},x
\right)
}.
\label{eq:app_action_importance_ratio}
\end{equation}

During optimization, the transition advantage is associated
only with the executable action produced at the corresponding
interaction step. Prompt tokens, environment observations,
retrieved content, padding tokens, and tokens belonging to
other interaction steps are excluded from this action-level
update to preserve precise step-wise credit assignment.

Using the rollout group
\(\mathcal{G}(x)=\{\tau_i\}_{i=1}^{N}\), the clipped
action-level objective is
\begin{equation}
\label{eq:final_objective} 
\begin{aligned}
  &\mathcal{J}_{\mathrm{TRCA}}(\theta) =\mathbb{E}
   \Bigl[\frac{1}{N}
    \sum_{i=1}^{N}\frac{1}{T_i}
    \sum_{t=1}^{T_i}
    \min \Big( \rho_{i,t}(\theta)A_{i,t}^{\text{TRCA}}, \\
    &\text{clip}(\rho_{i,t}(\theta), 1-\epsilon, 1+\epsilon) A_{i,t}^{\text{TRCA}} \Big) \Bigl] - \beta_{\text{KL}} \mathbb{D}_{\text{KL}}(\pi_\theta || \pi_{\text{ref}})
\end{aligned}
\end{equation}
Here, \(T_i\) is the number of environment interaction steps
in rollout \(i\), \(\epsilon\) is the PPO-style clipping
coefficient, and \(\beta_{\mathrm{KL}}\) controls
regularization toward the reference policy. Each term indexed
by \(t\) corresponds to one complete action-induced
transition and receives the trajectory-level outcome advantage and transition-wise rubric
signals.

Although the underlying language model generates an action
autoregressively, TRCA does not assign separate credit to
individual tokens within the same environment action. Instead,
all executable action tokens belonging to the same interaction
step are treated as realizing the same action \(a_{i,t}\) and
share the corresponding scalar transition advantage. This
action-level formulation is consistent with the objective
presented in the main text.

\subsection{Training Procedure}
\label{app:training_procedure}
Algorithm~\ref{alg:trca} summarizes the complete TRCA
training procedure. For each rollout group, TRCA evaluates
action-induced transitions using deterministic rubric
operators, constructs Foundational and Breakthrough Rubric
Rewards, and combines them with terminal outcomes to obtain
action-level advantages for policy optimization.
\begin{algorithm}[h]
\caption{Training TRCA}
\label{alg:trca}
\begin{algorithmic}[1]
\REQUIRE Policy \(\pi_{\theta}\), old policy
\(\pi_{\theta_{\mathrm{old}}}\), task distribution
\(p(\mathcal{X})\), benchmark-level operator libraries
\(\{\mathcal{T}_{b}\}\), deterministic task binders
\(\{\mathcal{B}_{b,m}\}\), benchmark adapters \(\{g_b\}\),
rollout group size \(N\), discount \(\gamma\), and mixing
coefficient \(\lambda\).
\FOR{each training iteration}
    \STATE Sample a batch of task instances \(x\sim p(\mathcal{X})\).
    \STATE Deterministically bind the operators in
    \(\mathcal{T}_{b}\) to each task \(x\), constructing
    \(\mathcal{R}(x)\).
    \STATE For each task instance, sample a rollout group of
    \(N\) trajectories using \(\pi_{\theta_{\mathrm{old}}}\).
    \STATE Collect terminal environment outcomes
    \(R(\tau_i)\) for all rollouts.
    \STATE Initialize rubric coverage \(C_{i,0,j}=0\) for all
    positive Evidence and Execution items.
    \FOR{each rollout \(\tau_i\)}
        \FOR{each transition \(\xi_{i,t}\)}
            \STATE Extract canonical observable facts using
            \(g_b(s_{i,t},a_{i,t},s_{i,t+1})\).
            \STATE Evaluate \(\mathcal{O}_{i,t,j}\) for all
            \(j\in\mathcal{R}(x)\) using the bound
            deterministic operators.
            \STATE Compute \(r^{\mathrm{F}}_{i,t}\) using
            Equation~\eqref{eq:app_foundational_reward}.
            \STATE Update coverage \(C_{i,t,j}\) using
            Equation~\eqref{eq:app_rubric_coverage}.
            \STATE Compute \(r^{\mathrm{B}}_{i,t}\) using
            Equation~\eqref{eq:app_breakthrough_expanded}.
            \STATE Compute \(r^{\mathrm{TRCA}}_{i,t}\) using
            Equation~\eqref{eq:app_trca_reward}.
        \ENDFOR
    \ENDFOR
    \STATE Construct comparable context groups
    \(\mathcal{G}^{\mathrm{S}}(\widetilde{s})\).
    \STATE Compute completion-aware returns \(R_{i,t}\) using
    Equation~\eqref{eq:app_completion_return}.
    \STATE Compute episode-relative advantages
    \(A^{\mathrm{E}}(\tau_i)\).
    \STATE Compute context-relative step advantages
    \(A^{\mathrm{S}}(a_{i,t})\).
    \STATE Construct final advantages
    \(A^{\mathrm{TRCA}}_{i,t}\).
    \STATE Optimize \(\pi_{\theta}\) with the clipped
    action-level objective.
    \STATE Update \(\pi_{\theta_{\mathrm{old}}}\leftarrow
    \pi_{\theta}\) before the next rollout collection phase.
\ENDFOR
\end{algorithmic}
\end{algorithm}

We next analyze the resulting credit signals in terms of
boundedness, novel-coverage preference, behavior in
zero-success groups, and conditional variance.

\subsection{Theoretical Properties of TRCA Credit Assignment}
\label{app:theoretical_properties}

This section establishes several theoretical properties of the transition-wise credit signals constructed by TRCA. We first show that the rubric rewards are bounded by the predefined category budgets and that Breakthrough Rubric Reward cannot accumulate through repeated satisfaction of the same rubric items. We then show that TRCA monotonically favors newly covered task-relevant conditions, retains discriminative step-level credit in zero-success rollout groups, and provides lower-variance local feedback than trajectory-outcome attribution throughout the policy optimization process.

\paragraph{Notation and assumptions.}
Let the positive rubric set be defined as
\begin{equation}
\mathcal{R}^{+}(x)
=
\mathcal{R}^{\mathrm{Evi}}(x)
\cup
\mathcal{R}^{\mathrm{Exec}}(x),
\label{eq:theory_positive_rubrics}
\end{equation}
which collects all Evidence and Execution items. For each
\(j\in\mathcal{R}^{c}(x)\), we define its normalized
category-specific weight as the following context 

\begin{equation}
w_j=\frac{r^c}{M_c}, \qquad M_c=\left|\mathcal{R}^{c}(x)\right|.
\label{eq:theory_item_weight}
\end{equation}
We assume that each category contains at least one valid rubric
item and that the transition evaluation operator is fully
deterministic given the corresponding observable
action-induced transition. These assumptions are fully
consistent with the rule-based transition evaluation adopted
by TRCA.

\paragraph{Proposition 1 (Rubric-budget boundedness and non-accumulation).}
Let
\begin{equation}
B_{+}=r^{\mathrm{Evi}}+r^{\mathrm{Exec}}, \qquad B_{-}=\left|r^{\mathrm{Inval}}\right|.
\label{eq:theory_reward_budgets}
\end{equation}
For every rollout $i$ and transition $t$, the reward components of TRCA satisfy
\begin{equation}
-B_{-}\leq r^F_{i,t}\leq B_{+}, \qquad 0\leq r^B_{i,t}\leq B_{+},
\label{eq:theory_component_bounds}
\end{equation}
and consequently
\begin{equation}
-(1-\lambda)B_{-}\leq r^{\mathrm{TRCA}}_{i,t}\leq B_{+}.
\label{eq:theory_trca_bound}
\end{equation}
Moreover, the cumulative Breakthrough Rubric Reward is bounded independently of the rollout horizon:
\begin{equation}
\sum_{t=1}^{T_i}r^B_{i,t}=\Phi_{i,T_i}\leq B_{+}.
\label{eq:theory_brr_accumulation}
\end{equation}

\textit{Proof of Proposition 1.}
For each positive category $c\in\{\mathrm{Evi},\mathrm{Exec}\}$, its aggregate contribution at transition $(i,t)$ satisfies the following constraints
\begin{equation}
0\leq \frac{r^c}{M_c}\sum_{j\in\mathcal{R}^{c}(x)}\mathcal{O}_{i,t,j}\leq r^c,
\label{eq:theory_positive_category_bound}
\end{equation}
because $\mathcal{O}_{i,t,j}\in\{0,1\}$ and $M_c=|\mathcal{R}^{c}(x)|$. Similarly, the aggregate Invalidity contribution satisfies
\begin{equation}
-B_{-}\leq-\frac{B_{-}}{N_{\mathrm{Inval}}}\sum_{j\in\mathcal{R}^{\mathrm{Inval}}(x)}\mathcal{O}_{i,t,j}\leq 0.
\label{eq:theory_invalidity_bound}
\end{equation}
Summing the category-wise contributions gives the bound on $r^F_{i,t}$. Breakthrough Rubric Reward contains only newly covered positive rubric items:
\begin{equation}
r^B_{i,t}=\sum_{c\in\{\mathrm{Evi},\mathrm{Exec}\}}\frac{r^c}{M_c}\sum_{j\in\mathcal{R}^{c}(x)}(1-C_{i,t-1,j})\mathcal{O}_{i,t,j}.
\label{eq:theory_brr_expanded}
\end{equation}
Every term in Equation~\eqref{eq:theory_brr_expanded} is non-negative, and the aggregate contribution of each category $c$ is at most $r^c$. Therefore, $0\leq r^B_{i,t}\leq B_{+}$.

Since
\begin{equation}
r^{\mathrm{TRCA}}_{i,t}=(1-\lambda)r^F_{i,t}+\lambda r^B_{i,t}, \qquad \lambda\in[0,1],
\end{equation}
the lower bound follows by combining the minimum possible Foundational Rubric Reward with zero breakthrough credit, while the upper bound follows because both $r^F_{i,t}$ and $r^B_{i,t}$ are upper bounded by $B_{+}$. Finally, since $r^B_{i,t}=\Phi_{i,t}-\Phi_{i,t-1}$ and $\Phi_{i,0}=0$, the cumulative reward telescopes:
\begin{equation}
\sum_{t=1}^{T_i}r^B_{i,t}=\sum_{t=1}^{T_i}\left(\Phi_{i,t}-\Phi_{i,t-1}\right)=\Phi_{i,T_i}.
\label{eq:theory_telescoping}
\end{equation}
Each positive rubric item contributes to $\Phi_{i,T_i}$ at most once, and the total contribution of category $c$ is at most $r^c$. Hence, $\Phi_{i,T_i}\leq B_{+}$. Thus, cumulative breakthrough credit is independent of rollout length and cannot grow through repeated satisfaction of previously covered rubric items.

\paragraph{Proposition 2 (Monotonicity with respect to novel rubric coverage).}
Consider two transition occurrences $u=(i,t)$ and
$v=(i',t')$ evaluated under the same task instruction and
grounded rubric set $\mathcal{R}(x)$. Let
$\mathcal{O}_{u,j}$ and $\mathcal{O}_{v,j}$ denote their
current rubric judgments, and define their prior coverage
states as
\begin{equation}
C^{-}_{u,j}=C_{i,t-1,j}, \qquad C^{-}_{v,j}=C_{i',t'-1,j}.
\label{eq:theory_prior_coverage}
\end{equation}
Suppose that the two transitions produce identical current rubric judgments:
\begin{equation}
\mathcal{O}_{u,j}=\mathcal{O}_{v,j}, \qquad \forall j\in\mathcal{R}(x).
\label{eq:theory_identical_judgments}
\end{equation}
Assume further that
\begin{equation}
C^{-}_{u,j}\leq C^{-}_{v,j}, \qquad \forall j\in\mathcal{R}^{+}(x),
\label{eq:theory_coverage_dominance}
\end{equation}
and that there exists at least one currently satisfied positive item $j^{\star}$ such that
\begin{equation}
\mathcal{O}_{u,j^{\star}}=\mathcal{O}_{v,j^{\star}}=1, \qquad C^{-}_{u,j^{\star}}=0, \qquad C^{-}_{v,j^{\star}}=1.
\label{eq:theory_strict_novel_item}
\end{equation}
Then
\begin{equation}
r^B_u>r^B_v.
\label{eq:theory_brr_monotonicity}
\end{equation}
Consequently, for every $\lambda>0$,
\begin{equation}
r^{\mathrm{TRCA}}_u>r^{\mathrm{TRCA}}_v.
\label{eq:theory_trca_monotonicity}
\end{equation}

\textit{Proof of Proposition 2.}
Because the current rubric judgments are identical, their signed item-level contributions are identical, and therefore
\begin{equation}
r^F_u=r^F_v.
\label{eq:theory_equal_frr}
\end{equation}
Using the expanded definition of Breakthrough Rubric Reward and $\mathcal{O}_{u,j}=\mathcal{O}_{v,j}$, we obtain
\begin{equation}
r^B_u-r^B_v=\sum_{c\in\{\mathrm{Evi},\mathrm{Exec}\}}\frac{r^c}{M_c}\sum_{j\in\mathcal{R}^{c}(x)}\left(C^{-}_{v,j}-C^{-}_{u,j}\right)\mathcal{O}_{u,j}.
\label{eq:theory_brr_difference}
\end{equation}
Every summand is non-negative by Equation~\eqref{eq:theory_coverage_dominance}. Moreover, the summand associated with $j^{\star}$ is strictly positive by Equation~\eqref{eq:theory_strict_novel_item}. Hence, $r^B_u-r^B_v>0$.

The difference between the combined TRCA rewards is
\begin{align}
r^{\mathrm{TRCA}}_u-r^{\mathrm{TRCA}}_v
&=(1-\lambda)(r^F_u-r^F_v)+\lambda(r^B_u-r^B_v) \nonumber\\
&=\lambda(r^B_u-r^B_v)>0.
\label{eq:theory_combined_difference}
\end{align}
Thus, among transitions providing the same immediate rubric evidence, TRCA strictly prefers a transition that newly covers a task-relevant condition over one that merely repeats an already covered condition.

\paragraph{Proposition 3 (Non-degenerate credit in zero-success rollout groups).}
Consider a rollout group $\mathcal{G}(x)=\{\tau_i\}_{i=1}^{N}$ whose terminal outcomes are identical:
\begin{equation}
R(\tau_1)=R(\tau_2)=\cdots=R(\tau_N).
\label{eq:theory_identical_outcomes}
\end{equation}
Then
\begin{equation}
A^E(\tau_i)=0, \qquad \forall i\in\{1,\ldots,N\}.
\label{eq:theory_zero_episode_advantage}
\end{equation}
In particular, this result holds when all sampled rollouts fail.

Now consider a context group $\mathcal{G}^{S}(\widetilde{s})$ containing at least two transitions. If the completion-aware returns in this group are not all identical, then the corresponding final TRCA advantages are not all zero. More precisely, there exist transitions $u,v\in\mathcal{G}^{S}(\widetilde{s})$ such that
\begin{equation}
A^{\mathrm{TRCA}}_u>0, \qquad A^{\mathrm{TRCA}}_v<0.
\label{eq:theory_non_degenerate_advantages}
\end{equation}

\textit{Proof of Proposition 3.}
Let $\mu_{\mathcal{G}}$ and $\sigma_{\mathcal{G}}$
denote the mean and standard deviation of the terminal
outcomes. Under Equation~\eqref{eq:theory_identical_outcomes},
every terminal outcome equals the group mean and
$\sigma_{\mathcal{G}}=0$. By the normalization convention in
Equation~\eqref{eq:app_episode_advantage},
\begin{equation}
A^E(\tau_i)=0.
\label{eq:theory_episode_advantage_proof}
\end{equation}

Let
\begin{equation}
\mathcal{H}=\left\{R_{j,k}\mid(j,k)\in\mathcal{G}^{S}(\widetilde{s})\right\}
\label{eq:theory_return_set}
\end{equation}
denote the set of completion-aware returns in the context group, and let $\mu_{\mathcal{H}}$ and $\sigma_{\mathcal{H}}$ denote its mean and standard deviation. Since the returns are not all identical,
\begin{equation}
\min\mathcal{H}<\mu_{\mathcal{H}}<\max\mathcal{H}.
\label{eq:theory_mean_between_extrema}
\end{equation}
Hence, there exist transitions $u$ and $v$ satisfying $R_u>\mu_{\mathcal{H}}$ and $R_v<\mu_{\mathcal{H}}$. Since $\sigma_{\mathcal{H}}+\epsilon_{\mathrm{norm}}>0$, normalization preserves their signs:
\begin{equation}
A^S_u=\frac{R_u-\mu_{\mathcal{H}}}{\sigma_{\mathcal{H}}+\epsilon_{\mathrm{norm}}}>0, \qquad A^S_v=\frac{R_v-\mu_{\mathcal{H}}}{\sigma_{\mathcal{H}}+\epsilon_{\mathrm{norm}}}<0.
\label{eq:theory_step_advantage_signs}
\end{equation}

The final TRCA advantage is $A^{\mathrm{TRCA}}_{i,t}=A^E(\tau_i)+A^S(a_{i,t})$. Since $A^E(\tau_i)=0$ for every rollout in the group,
\begin{equation}
A^{\mathrm{TRCA}}_u=A^S_u>0, \qquad A^{\mathrm{TRCA}}_v=A^S_v<0.
\end{equation}
Therefore, whenever rubric-enhanced completion-aware returns distinguish actions arising from the same decision context, TRCA provides discriminative positive and negative credit even when the rollout group contains no successful trajectory throughout early-stage policy optimization and exploration under otherwise success-sparse training conditions.

\paragraph{Proposition 4 (Conditional variance reduction of transition-wise feedback).}
Consider rollouts sampled from a fixed policy $\pi$. Define the augmented local transition information as
\begin{equation}
Z_{i,t}=\left(x,t,\xi_{i,t},\mathbf{C}_{i,t-1}\right),
\label{eq:theory_augmented_transition}
\end{equation}
where
\begin{equation}
\begin{aligned}
\xi_{i,t}
&= (s_{i,t}, a_{i,t}, s_{i,t+1}), \\
\mathbf{C}_{i,t-1}
&= \{ C_{i,t-1,j} \}_{j \in \mathcal{R}^{+}(x)} .
\end{aligned}
\label{eq:theory_transition_and_coverage}
\end{equation}

Let the local TRCA feedback be $X^{\mathrm{TRCA}}_{i,t}=r^{\mathrm{TRCA}}_{i,t}$, and let the trajectory-outcome feedback be $X^E_{i,t}=R(\tau_i)$, where $R(\tau_i)\in\{0,1\}$. Under deterministic rule-based transition evaluation, for every $z$ visited with positive probability,
\begin{equation}
\operatorname{Var}\left(X^{\mathrm{TRCA}}_{i,t}\mid Z_{i,t}=z\right)\leq\operatorname{Var}\left(X^E_{i,t}\mid Z_{i,t}=z\right).
\label{eq:theory_variance_inequality}
\end{equation}
The inequality is strict whenever both successful and failed continuations occur with positive probability conditioned on $Z_{i,t}=z$.

\textit{Proof of Proposition 4.}
Under deterministic transition evaluation, each rubric judgment is a deterministic function of the observable transition:
\begin{equation}
\mathcal{O}_{i,t,j}
=
h_{b,m}^{c}\left(
\omega_{b,m}(x),
g_b(\xi_{i,t})
\right),
\qquad
j=(h_{b,m}^{c},\omega_{b,m}(x)).
\label{eq:theory_deterministic_operator}
\end{equation}
Consequently, Foundational Rubric Reward $r^F_{i,t}=\sum_{j\in\mathcal{R}(x)}q^i_{t,j}$ is fixed once $Z_{i,t}=z$ is given. Since $Z_{i,t}$ also contains the prior coverage state $\mathbf{C}_{i,t-1}$, Breakthrough Rubric Reward
\begin{equation}
r^B_{i,t}=\sum_{c\in\{\mathrm{Evi},\mathrm{Exec}\}}\frac{r^c}{M_c}\sum_{j\in\mathcal{R}^{c}(x)}(1-C_{i,t-1,j})\mathcal{O}_{i,t,j}
\end{equation}
is also fixed conditioned on $Z_{i,t}=z$. Therefore, $r^{\mathrm{TRCA}}_{i,t}$ is a deterministic function of $Z_{i,t}$, yielding
\begin{equation}
\operatorname{Var}\left(X^{\mathrm{TRCA}}_{i,t}\mid Z_{i,t}=z\right)=0.
\label{eq:theory_zero_local_variance}
\end{equation}

Define the conditional success probability as
\begin{equation}
p_z=\Pr\left(R(\tau_i)=1\mid Z_{i,t}=z\right).
\label{eq:theory_conditional_success}
\end{equation}
Since $R(\tau_i)$ is binary, its conditional first and second moments are both equal to $p_z$. Therefore,
\begin{equation}
\operatorname{Var}\left(X^E_{i,t}\mid Z_{i,t}=z\right)=p_z-p_z^2=p_z(1-p_z)\geq 0.
\label{eq:theory_outcome_variance}
\end{equation}
Combining Equations~\eqref{eq:theory_zero_local_variance} and~\eqref{eq:theory_outcome_variance} proves Equation~\eqref{eq:theory_variance_inequality}.

If both successful and failed continuations have positive conditional probability, then $0<p_z<1$ and $p_z(1-p_z)>0$. It follows that
\begin{equation}
\operatorname{Var}\left(X^{\mathrm{TRCA}}_{i,t}\mid Z_{i,t}=z\right)<\operatorname{Var}\left(X^E_{i,t}\mid Z_{i,t}=z\right).
\end{equation}
Thus, after the observable transition and prior rubric coverage are fixed, TRCA provides deterministic local feedback, whereas trajectory-outcome attribution remains affected by subsequent policy decisions.

\paragraph{Scope of the analysis.}
The above results characterize the boundedness, novel-coverage preference, non-degeneracy, and conditional variance of the transition-level credit signals constructed by TRCA. Proposition~4 concerns the local transition feedback before context-relative normalization. It does not imply that the complete normalized advantage or the resulting policy-gradient estimator necessarily has zero or uniformly lower variance. The analysis also does not assume that TRCA is an unbiased estimator of an unknown optimal advantage function or claim policy invariance under reward shaping within the present theoretical scope.
\section{Diagnostic Study Protocol}
\label{app:diagnostic_study}

\subsection{Rollout Collection}

The diagnostic study is conducted on ALFWorld and WebShop
using Qwen2.5-1.5B-Instruct during the early stage of
reinforcement learning. Each diagnostic batch contains 16 task
instances with a rollout group size of 8, yielding 128
rollouts per batch. A failed rollout is defined as a rollout
with terminal outcome \(R(\tau_i)=0\). A success-free group
is a task-conditioned rollout group in which none of the
sampled rollouts achieves terminal success. The reported percentages are aggregate averages across multiple diagnostic batches and benchmarks rather than counts computed from a single 128-rollout batch under one experimental configuration.

The diagnostic analysis reports three aggregate statistics:
\(96.5\%\) of sampled rollouts fail to achieve terminal
success, \(85.6\%\) of task-conditioned rollout groups
contain no successful trajectory, and \(72.2\%\) of actions
in failed rollouts retain diagnostically useful transition
signals.

\subsection{Transition Annotation}

The annotation unit is an action-induced transition
\(\xi_{i,t}=(s_{i,t},a_{i,t},s_{i,t+1})\), rather than a
complete trajectory. We annotate the collected transitions
using human assessment together with GPT-4o as the frontier
LLM judge.

The diagnostic labels distinguish \textsc{Evidence},
\textsc{Execution}, \textsc{Validity}, and \textsc{Other}.
\textsc{Evidence} identifies transitions that reveal
task-relevant information. \textsc{Execution} identifies
transitions that complete a substantive task-required
operation or intermediate condition. \textsc{Validity}
identifies transitions whose actions are successfully parsed
and executed, even when they do not immediately complete a
subgoal. \textsc{Other} denotes transitions without a useful
signal under the diagnostic annotation protocol.

This LLM-assisted annotation is used only for the diagnostic
analysis reported in Figure~\ref{fig:Motivation} and is never
used to construct transition rewards during TRCA training.
During policy optimization, all rubric judgments are computed
by the deterministic benchmark-level operators described in
Appendix~\ref{app:rule_based_evaluation}, without any additional online model inference.

\subsection{Calculation of Reported Ratios}
\label{app:diagnostic_ratios}

The reported ratios use different denominators. We define
\begin{equation}
\begin{aligned}
\mathrm{FailureRate}
&= N_{\mathrm{fail}} / N_{\mathrm{rollout}},\\
\mathrm{SuccessFreeGroupRate}
&= N_{\mathrm{sf\_group}} / N_{\mathrm{group}},\\
\mathrm{UsefulTransitionRate}
&= N_{\mathrm{useful}}^{\mathrm{fail}} /
N_{\mathrm{action}}^{\mathrm{fail}} .
\end{aligned}
\label{eq:app_diagnostic_ratios}
\end{equation}
In the main text, these correspond to \(96.5\%\), \(85.6\%\),
and \(72.2\%\), respectively. The first two are rollout-level
or group-level statistics. The third is an action-level
statistic computed only over actions in failed rollouts. A
useful action is one annotated as providing Evidence,
Execution, or Validity under the diagnostic annotation
categories.

\subsection{Annotation Categories and Quality Control}
\label{app:diagnostic_quality_control}

Evidence, Execution, Validity, and Other serve only the
diagnostic study. Validity is complementary to the Invalidity
rubric used during TRCA reward construction: a valid action is
successfully parsed and executed, whereas Invalidity captures
malformed, inadmissible, rejected, or otherwise unexecutable
actions and contributes negative reward.

Quality control requires separating diagnostic annotation from
training-time reward evaluation. Human annotation and the
frontier LLM judge are used to analyze the motivating
failure-scarce regime, while TRCA training does not use an
online LLM judge at any stage of policy optimization.

\section{Experimental Details}
\label{app:experimental_details}

\subsection{Models and Rollout Configuration}
\label{app:model_configuration}

\begin{figure*}[t]
    \centering
    \includegraphics[width=\textwidth]{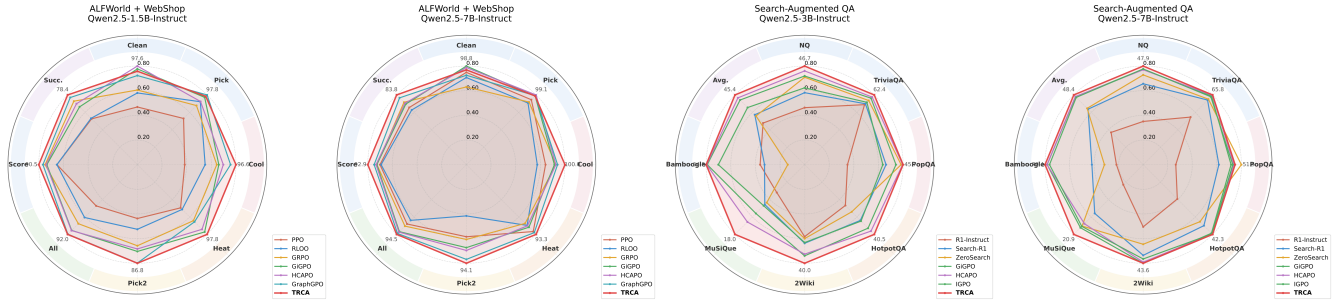}
    \caption{Performance profiles of TRCA and representative baselines across
model scales. The first two panels summarize ALFWorld subtask
success rates and WebShop task score and success rate, while the
last two panels summarize performance on seven search-augmented QA
benchmarks and their average. Values are scaled to $[0,1]$ for
visualization.}
    \label{fig:radar}
\end{figure*}

For ALFWorld and WebShop, the experiments use
Qwen2.5-1.5B-Instruct and Qwen2.5-7B-Instruct as policy
models. For SearchQA, we use Qwen2.5-3B-Instruct and
Qwen2.5-7B-Instruct. All group-based methods use rollout
group size \(N=8\). The agent emits reasoning within
\texttt{<think>} tags and its executable action within
\texttt{<action>} tags. Unless otherwise specified, the mixing
coefficient between Foundational Rubric Reward and
Breakthrough Rubric Reward is \(\lambda=0.8\). For the ALFWorld training configuration, the maximum number of
interaction steps is 25, the rollout temperature is 1.0, and
the maximum response length is 512 tokens.
To complement the numerical results reported in the main tables,
Figure~\ref{fig:radar} provides an overall view of the
performance profiles of TRCA and representative baselines across
benchmarks and model scales. The first two panels summarize
ALFWorld subtask performance together with the WebShop task score
and success rate, while the last two panels summarize results on
the seven SearchQA benchmarks and their average. TRCA maintains
a consistently strong and balanced profile, indicating that its
improvements are not concentrated on a single task category or
model scale.

\subsection{Training Hyperparameters}
\label{app:training_hyperparameters}

Table~\ref{tab:app_hyperparameters} summarizes the core
hyperparameters used for reinforcement learning. Engineering
parameters related to distributed execution, logging, and
checkpoint management are omitted for clarity.

\begin{table}[t]
\centering
\small
\caption{Core reinforcement-learning hyperparameters used in the experiments.}
\label{tab:app_hyperparameters}
\begin{tabular}{lc}
\toprule
\textbf{Hyperparameter} & \textbf{Value} \\
\midrule
Learning rate & \(8\times10^{-7}\) \\
Rollout group size \(N\) & 8 \\
PPO mini-batch size & 256 \\
PPO micro-batch size per GPU & 32 \\
Discount factor \(\gamma\) & 0.95 \\
PPO clipping coefficient (lower) & 0.20 \\
PPO clipping coefficient (upper) & 0.28 \\
KL regularization coefficient & 0.012 \\
Entropy coefficient & 0.001 \\
Rollout temperature & 1.0 \\
Maximum response length & 512 \\
TRCA mixing coefficient \(\lambda\) & 0.8 \\
\bottomrule
\end{tabular}
\end{table}

\subsection{Benchmark-Specific Context Grouping}
\label{app:context_grouping_details}

Context grouping uses observable, benchmark-specific state
information to compare actions taken under similar decision
contexts. It is used for computing the context-relative step
advantage and does not depend on successful terminal states.

\begin{table*}[t]
\centering
\small
\caption{Observable context information used to define
comparable decision contexts.}
\label{tab:app_context_fields}
\begin{tabular}{p{0.18\linewidth}p{0.72\linewidth}}
\toprule
Benchmark & Comparable-context information \\
\midrule
ALFWorld &
Task predicates, object state, inventory, object location, and
interaction stage. \\
WebShop &
Page type, product identity, selected options, product
attributes, and navigation stage. \\
SearchQA &
Tool type, retrieval stage, query state, evidence stage, and
answer-submission stage. \\
\bottomrule
\end{tabular}
\end{table*}

The grouping is structural rather than purely lexical: exact
textual equality is not required when the environment exposes
equivalent structured state fields. If a context group has
fewer than two comparable transitions or has zero variance in
completion-aware returns, the corresponding
\(A^{\mathrm{S}}(a_{i,t})\) is set to zero. Context grouping is implemented using deterministic structural fields exposed by each environment, without embedding-based clustering, learned state encoders, or additional LLM inference.

\subsection{Datasets and Evaluation Protocols}
\label{app:datasets_evaluation}

ALFWorld is evaluated with success rate. The main table
reports the five ALFWorld task categories Clean, Pick, Cool,
Heat, and Pick2, together with the overall success rate.
WebShop is evaluated with both task score and success rate:
task score captures partial attribute satisfaction, while
success rate captures exact task completion.

SearchQA includes seven datasets: Natural Questions (NQ),
TriviaQA, PopQA, HotpotQA, 2WikiMultiHopQA (2Wiki), MuSiQue,
and Bamboogle. NQ and HotpotQA are treated as in-domain
benchmarks; TriviaQA, PopQA, 2Wiki, MuSiQue, and Bamboogle
are treated as out-of-domain benchmarks. SearchQA reports performance using strict binary normalized
Exact Match, counting an example as successful only when its
normalized final answer exactly matches one of the provided
ground-truth aliases.

\subsection{Baseline Configuration and Comparison Protocol}
\label{app:baseline_protocol}

The main experiments compare TRCA with closed-source
prompting baselines, prompting-based agents, and RL training
methods. The closed-source baselines are GPT-4o and
Gemini-2.5-Pro. Prompting-based agents include Qwen2.5,
ReAct, and Reflexion. RL baselines for ALFWorld and WebShop
include PPO, RLOO, GRPO, GiGPO, HCAPO, and GraphGPO. For
SearchQA, the compared methods are R1-Instruct, Search-R1,
ZeroSearch, StepSearch, GiGPO, HCAPO, IGPO, and TRCA. For SearchQA, we follow the training and evaluation protocol
of GiGPO. For ALFWorld and WebShop, we evaluate all methods
under the Qwen2.5-1.5B-Instruct and Qwen2.5-7B-Instruct
settings.

\subsection{Sample-Efficiency Evaluation Protocol}
\label{app:sample_efficiency_protocol}

Figure~\ref{fig:placeholder} reports sample efficiency on the
ALFWorld \textit{Pick\_two} and \textit{Pick\_cool} subtasks
using Qwen2.5-1.5B-Instruct. The counting unit is a sampled
trajectory, i.e., a complete rollout. In general,
\begin{equation}
N_{\mathrm{traj}}
=
N_{\mathrm{iterations}}
\times
N_{\mathrm{tasks/iteration}}
\times
N_{\mathrm{rollouts/task}}.
\label{eq:app_trajectory_budget}
\end{equation}
We count each completed rollout as one sampled trajectory.
The reported aggregate budgets are 1.92K, 3.84K, and 6.40K
sampled trajectories over the complete evaluation process.

\begin{figure}[t]
    \centering

    \includegraphics[width=0.92\columnwidth]{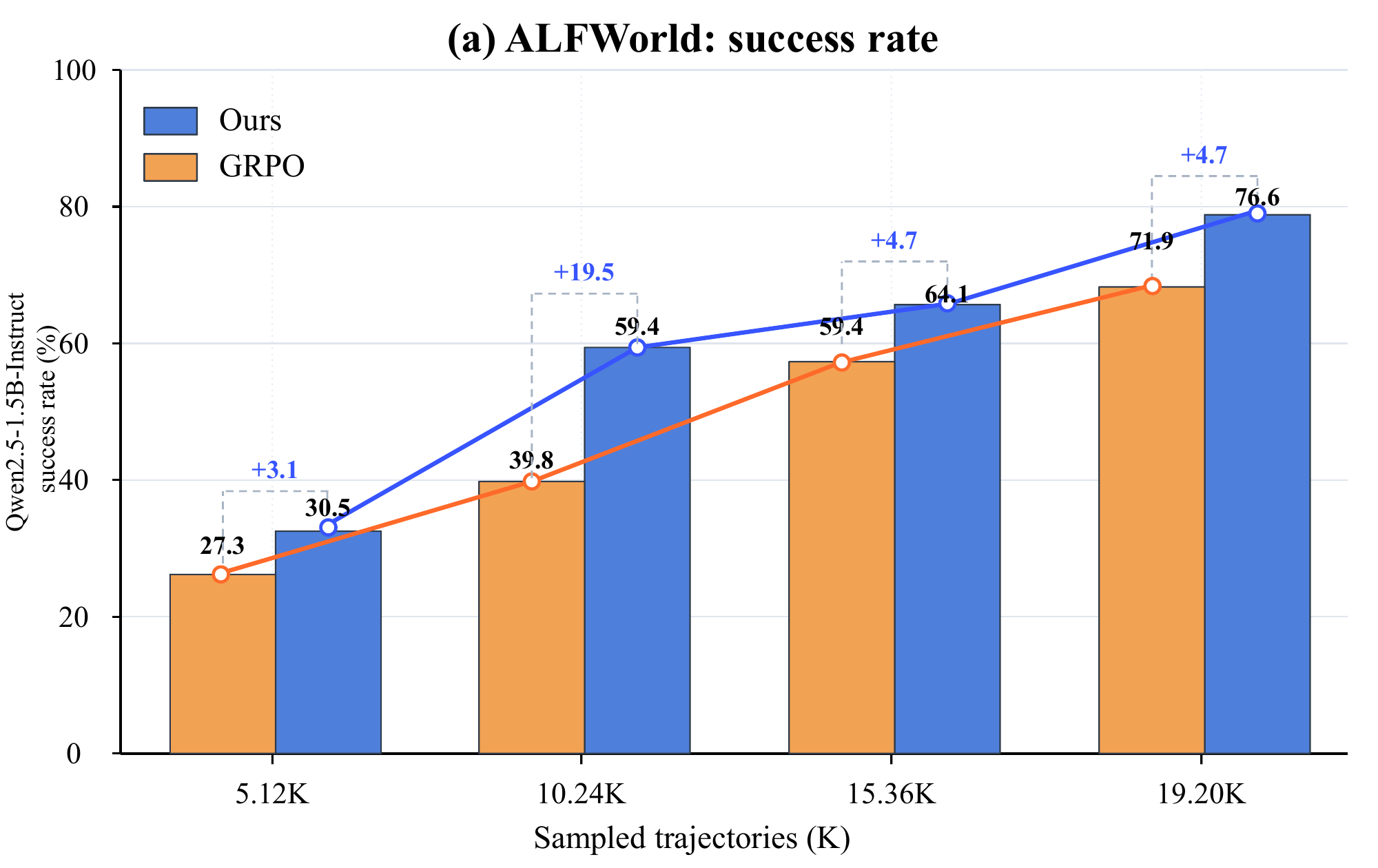}

    \vspace{-0.25cm}

    \includegraphics[width=0.92\columnwidth]{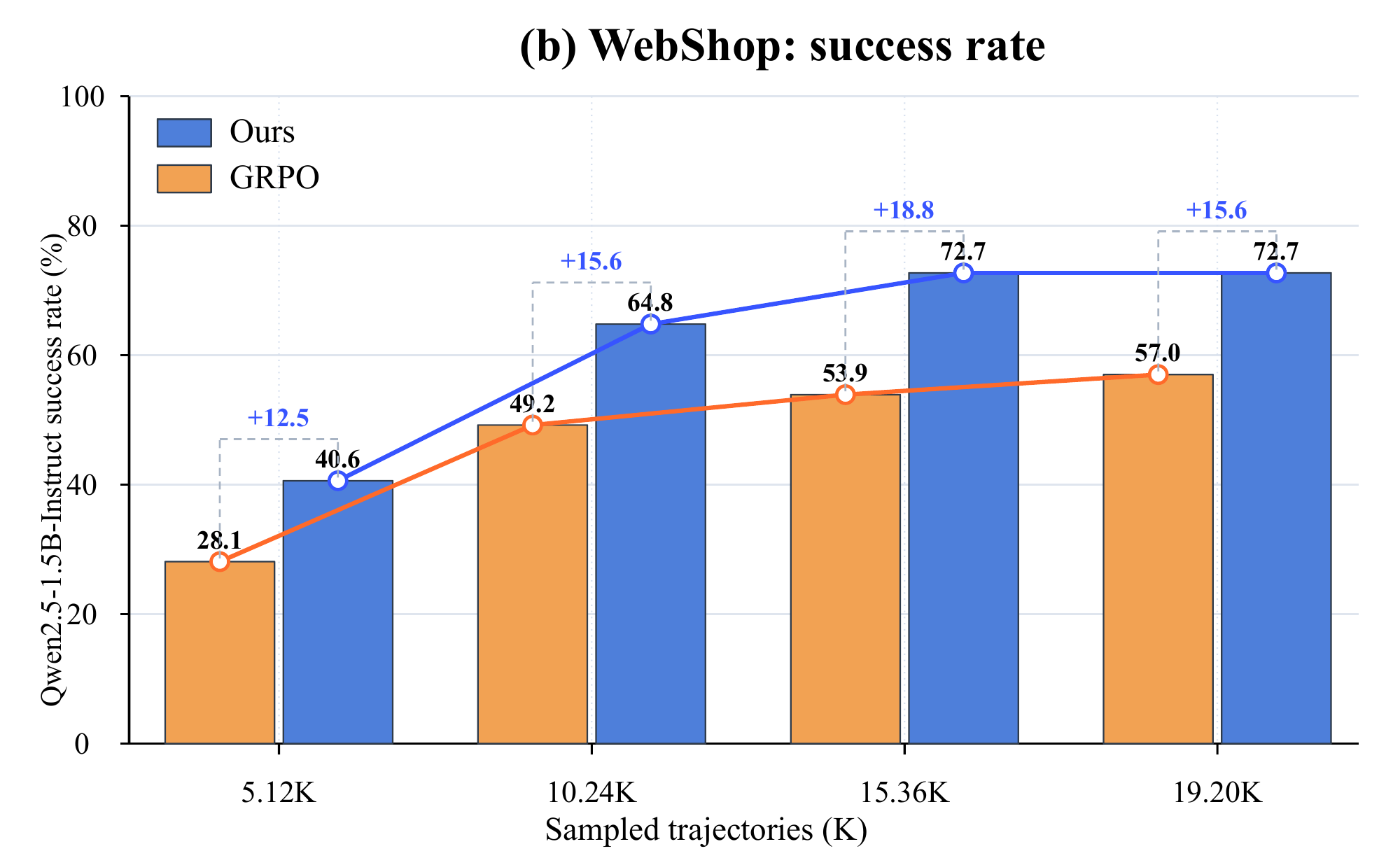}

    \vspace{-0.15cm}
    \caption{Sample-efficiency comparison between TRCA and GRPO on ALFWorld and WebShop using Qwen2.5-1.5B-Instruct. TRCA consistently outperforms GRPO across sampled-trajectory budgets, showing stronger sample efficiency under identical interaction and optimization settings.}
    \label{fig:sample_efficiency}

    \vspace{-0.25cm}
\end{figure}

In addition to the subtask-level comparison reported in the main
text, Figure~\ref{fig:sample_efficiency} presents the
aggregate sample-efficiency results on ALFWorld and WebShop under
matched rollout budgets. TRCA consistently achieves higher success
rates than GRPO across all reported budgets, with gains of up to
$19.5$ percentage points on ALFWorld and $18.8$ percentage points
on WebShop. The performance gap emerges early and remains visible
as the sampling budget increases, showing that transition-wise
rubric rewards extract more effective supervision from the same
amount of agent--environment interaction. At 1.92K sampled trajectories, TRCA improves \textit{Pick\_two}
success rate from 11.5\% to 34.6\% and \textit{Pick\_cool}
success rate from 15.0\% to 24.0\% relative to GRPO under the
same budget. On \textit{Pick\_two}, TRCA reaches 69.2\%
success rate with 3.84K trajectories, exceeding the 46.7\%
achieved by GRPO with 6.40K trajectories. On
\textit{Pick\_cool}, TRCA reaches 53.8\% with 3.84K
trajectories, approaching the 57.7\% GRPO result at 6.40K
trajectories. The statement that TRCA uses 40\% fewer samples
follows from
\[
1-\frac{3.84\mathrm{K}}{6.40\mathrm{K}}=0.40.
\]

\subsection{Random Seeds and Reporting}
\label{app:reporting_protocol}

\begin{figure}[t]
    \centering
    \includegraphics[width=\linewidth]{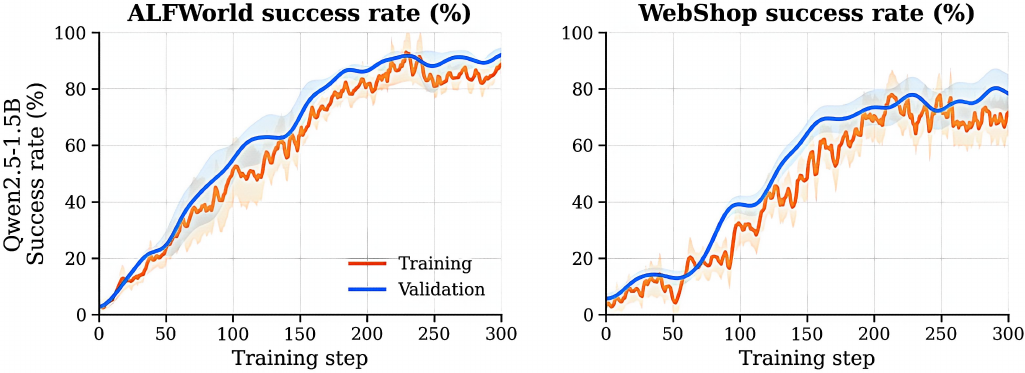}
    \caption{Training and validation success rates of TRCA on ALFWorld and WebShop.}
    \label{fig:trca_train_validation}
    \vspace{-0.2cm}
\end{figure}

Results on ALFWorld and WebShop are averaged over three
random seeds. Table~\ref{tab:main_results} shows the standard deviation of the results from three runs. 
Figure~\ref{fig:trca_train_validation} further reports the
training and validation success rates of TRCA on ALFWorld and
WebShop. In both environments, the validation curves generally
follow the improvement of the corresponding training curves,
without pronounced divergence during optimization. This agreement provides a qualitative check that observed training gains are reflected in evaluation performance.

Experiments are conducted on four NVIDIA A800 80GB GPUs.
The method itself does not require online LLM-judge inference
for rubric evaluation during policy training. Benchmark-level
operator libraries are constructed once offline; deterministic
task binders then instantiate task-grounded rubric items, and
transition judgments are computed from observable environment
feedback without additional LLM calls. This distinguishes
TRCA training from the diagnostic study, where human
annotation and a frontier LLM judge are used only to analyze
the motivating regime.

\begin{figure}[t]
    \centering
    \includegraphics[width=\linewidth]{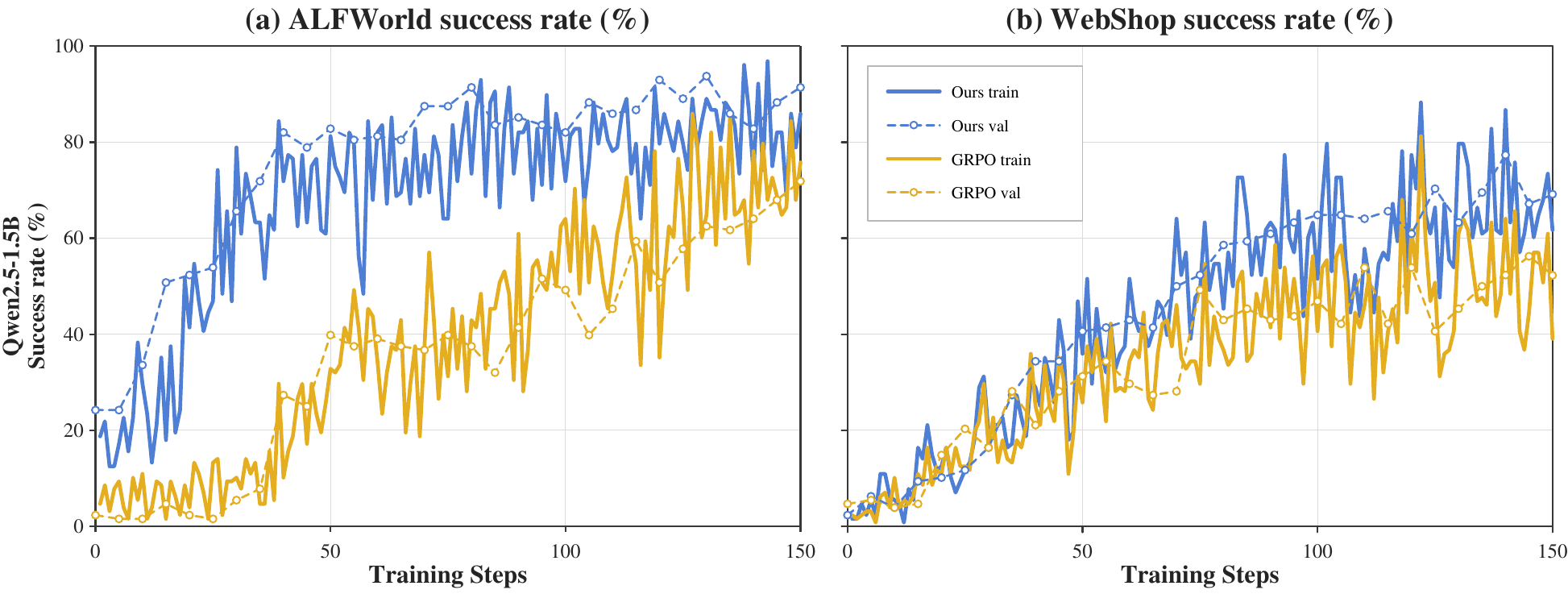}
    \caption{Training and validation success rates of TRCA and
GRPO on WebShop and ALFWorld. TRCA converges faster and
achieves stronger performance, on ALFWorld.}
    \label{fig:trca_grpo_train_validation}
    \vspace{-0.5cm}
\end{figure}

To examine whether the optimization gains transfer from the
sampled training tasks to held-out evaluation tasks,
Figure~\ref{fig:trca_grpo_train_validation} compares the
training and validation success-rate trajectories of TRCA and
GRPO throughout optimization. In Figure~\ref{fig:trca_grpo_train_validation}(a),
TRCA exhibits a pronounced early-stage advantage on ALFWorld.
Its training and validation performance rises rapidly within
the first several dozen updates, reaching a high-success
regime substantially earlier than GRPO. By contrast, GRPO
remains near the low-performance region during the initial
stage and improves only gradually afterward. Although the
training curves of both methods contain stochastic
fluctuations, the validation trajectory of TRCA remains
consistently above that of GRPO over most of the training
process, indicating both faster convergence and a higher
held-out performance level. The gap persists into the later
stage, where TRCA maintains success rates around the upper
performance range, whereas GRPO continues to improve more
slowly and converges at a lower level.

Figure~\ref{fig:trca_grpo_train_validation}(b) shows a similar,
though more gradual, pattern on WebShop. The two methods begin
from comparable low success rates and improve steadily during
early optimization. After approximately the first third of
training, however, TRCA begins to establish a persistent
advantage in both the training and validation curves. This
advantage becomes clearer during the middle and late stages,
where TRCA generally reaches higher success rates and retains
a stronger validation trajectory despite occasional
step-level fluctuations. In both panels, the validation curves
closely follow the overall trends of their corresponding
training curves without persistent divergence. These results
suggest that TRCA not only learns more rapidly from the
sampled rollouts, but also achieves stronger final success
rates and more stable behavior than GRPO.
The consistent ordering between training and held-out
performance further indicates that the optimization gains are
not confined to the sampled training instances, but transfer
reliably to unseen evaluation tasks.


\section{Rubric Generation Prompt and Complete Libraries}
\label{app:rubric_prompt}


Before policy training, an LLM is invoked once for each
benchmark to construct a shared rubric operator library.
The prompt specifies five deterministic Boolean operators
for each of three categories: \textsc{Evidence}, which detects
newly revealed task-relevant information; \textsc{Invalidity},
which detects malformed, rejected, repeated, or otherwise
unexecutable actions; and \textsc{Execution}, which detects
accepted task-required operations, achieved subgoals, or
satisfied intermediate conditions. Each operator specifies
the observable facts it reads and an exact Boolean activation
condition based only on the task instruction and the
action-induced transition. Missing, ambiguous, or unsupported
observable evidence produces a zero judgment. Successful
trajectories, terminal outcomes, future states, and
ground-truth answers are not used to evaluate intermediate
transitions. The resulting operator library is shared across all task
instances within the same benchmark, while task-specific
entities, attributes, relations, and constraints are supplied
through deterministic parameter binding. Once these bindings
are established, the grounded operators are applied
deterministically to every transition during rollout
collection and policy optimization without further LLM calls.
This design also keeps the overall evaluation procedure
consistent and reproducible across task instances.

\begin{figure*}[!t]
    \centering

    \includegraphics[width=0.95\linewidth]{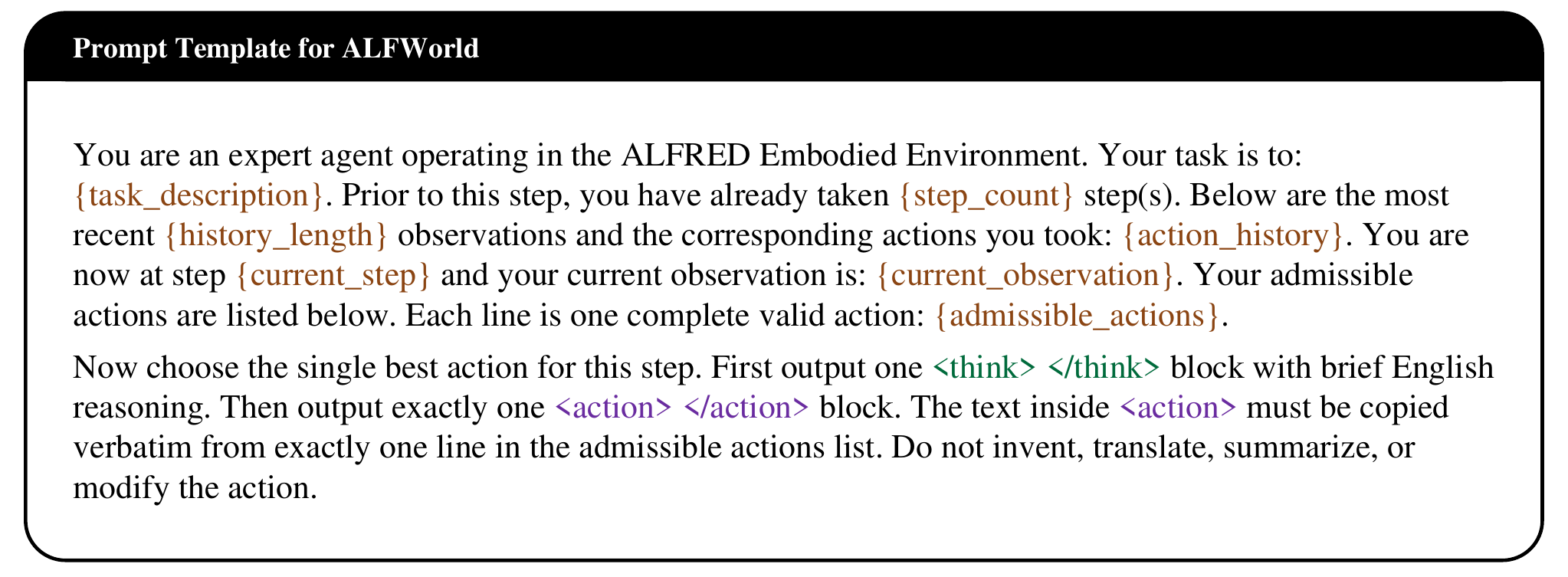}
    \vspace{-0.2cm}

    \includegraphics[width=0.95\linewidth]{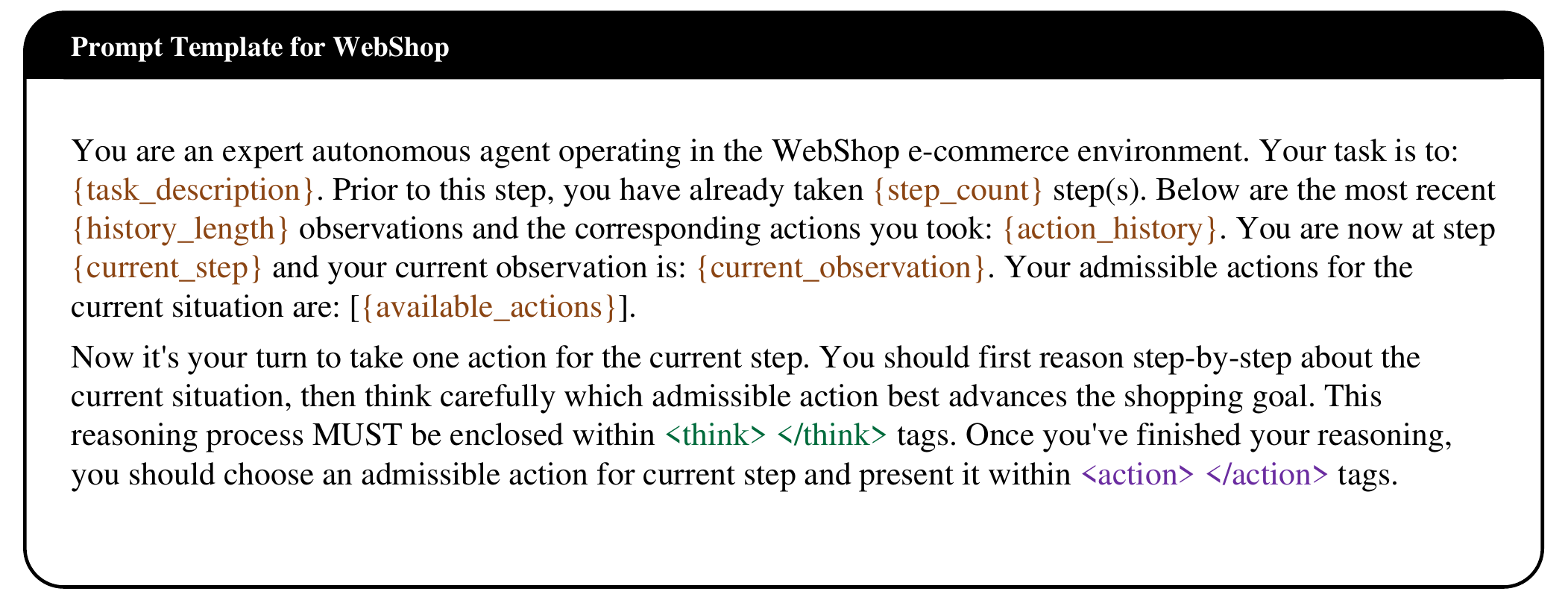}
    \vspace{-0.2cm}

    \includegraphics[width=0.95\linewidth]{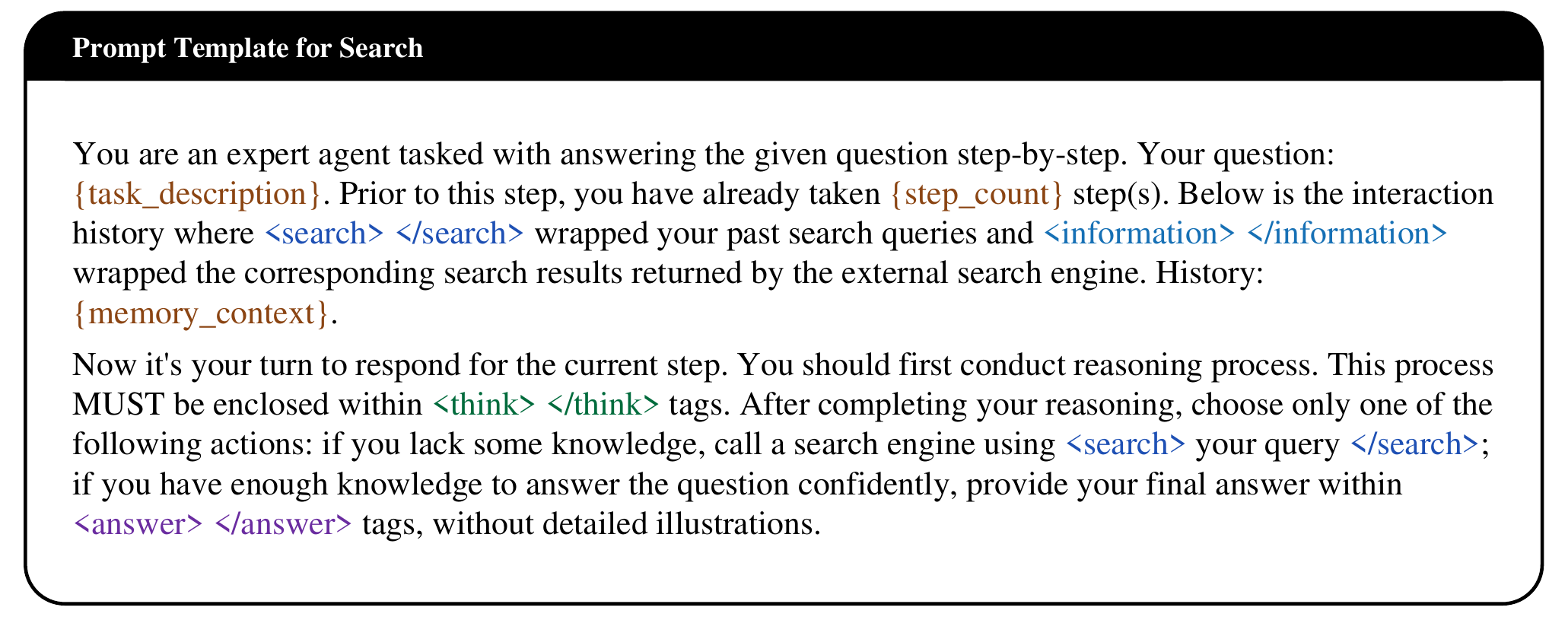}

    \caption{Prompt templates used for ALFWorld, WebShop, and search-augmented QA agents.}
    \label{fig:prompt_templates}
\end{figure*}

\clearpage
\FloatBarrier
\clearpage
\includepdf[pages=-,scale=0.95,pagecommand={}]{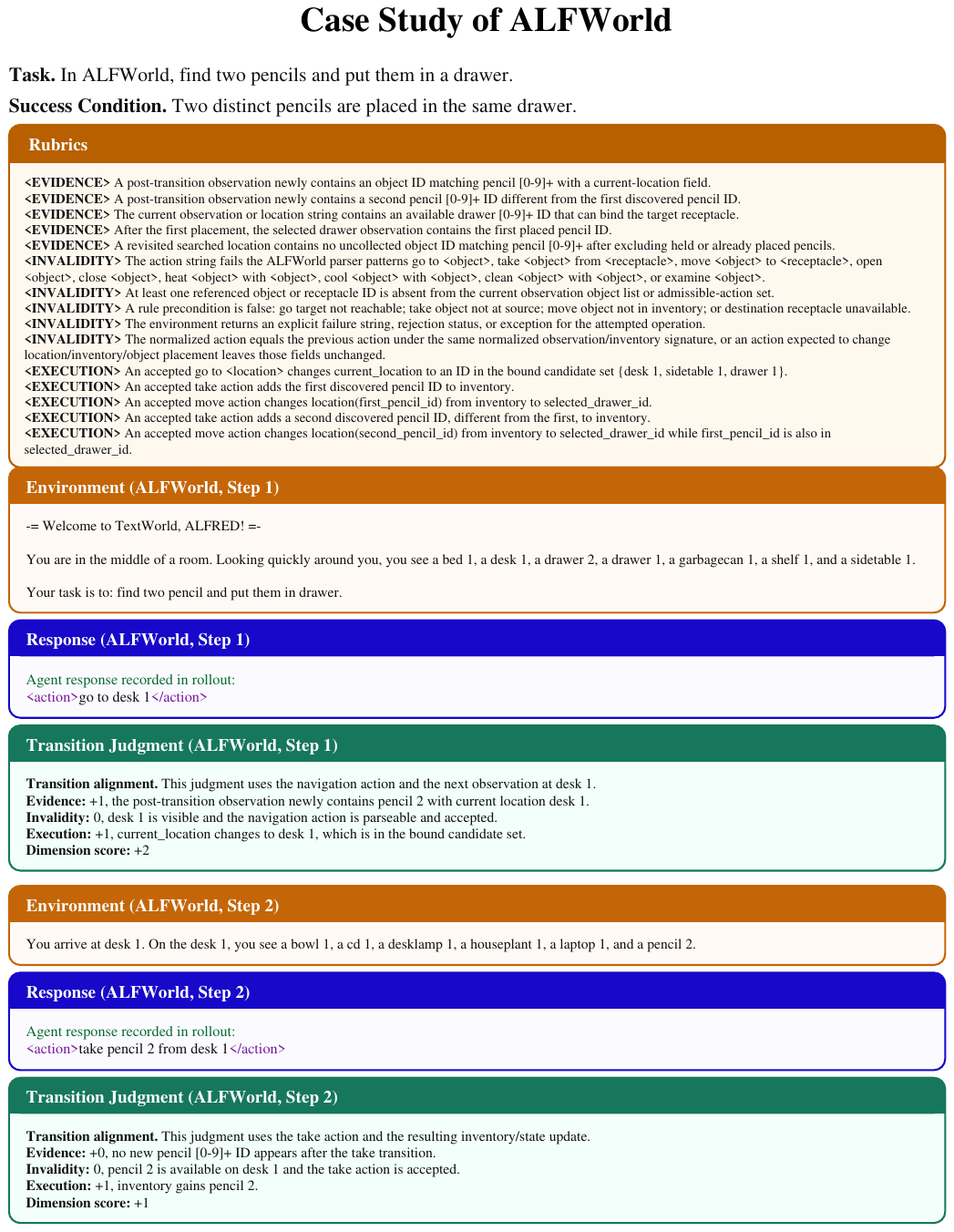}
\includepdf[pages=-,scale=0.95,pagecommand={}]{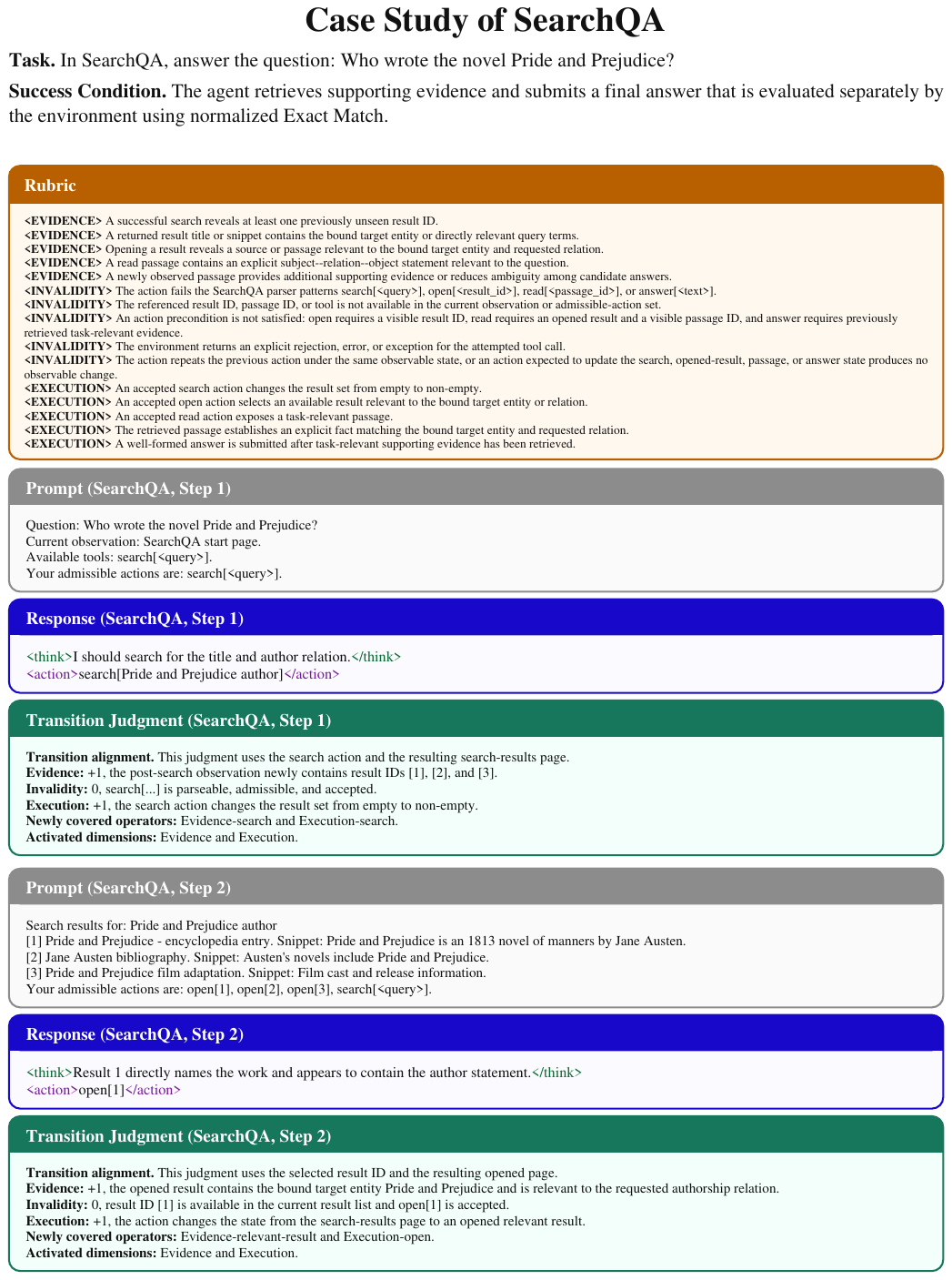}
\includepdf[pages=-,scale=0.95,pagecommand={}]{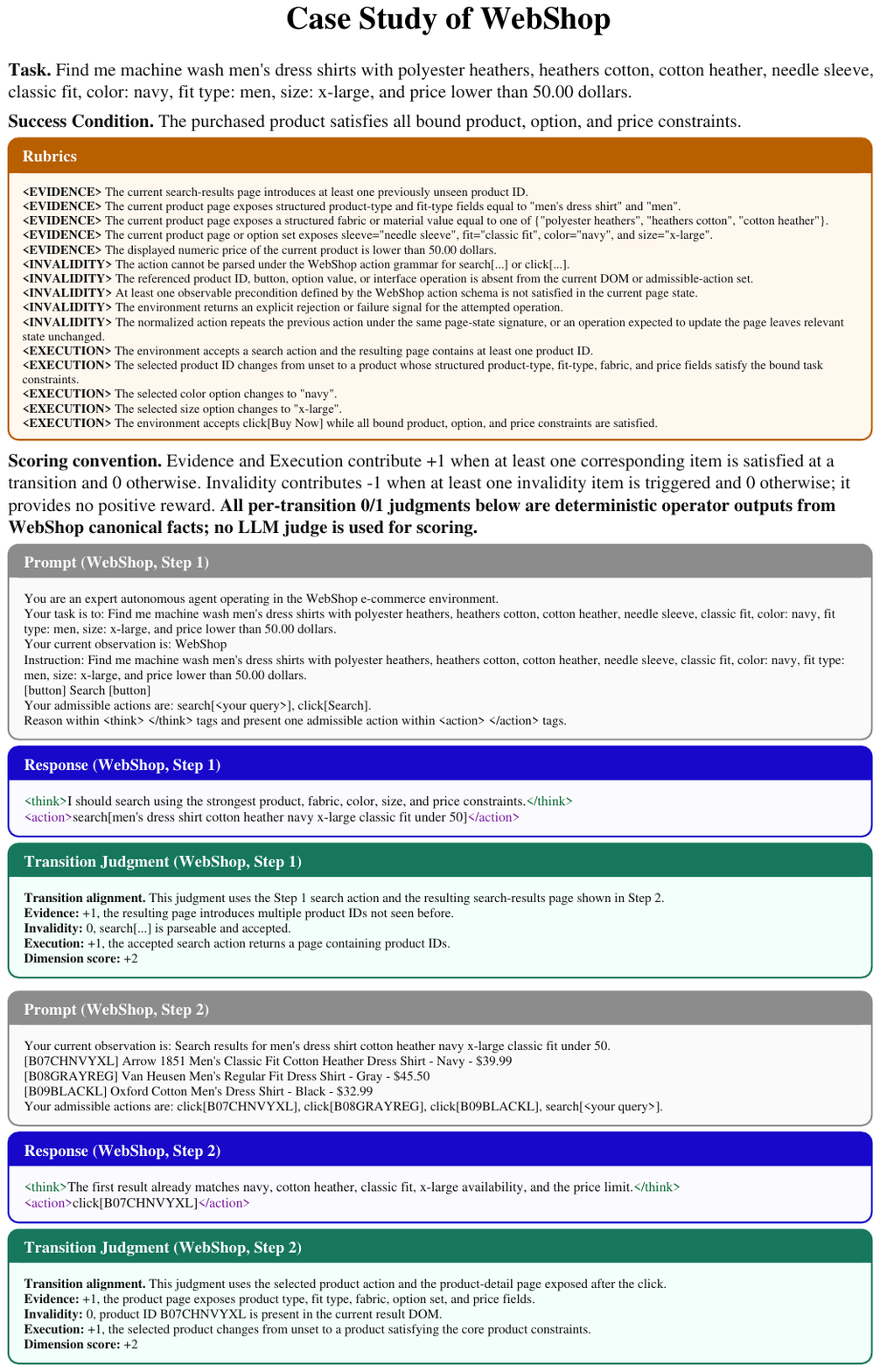}

\begin{figure*}[t]
    \centering
    \includegraphics[width=1.0\textwidth]{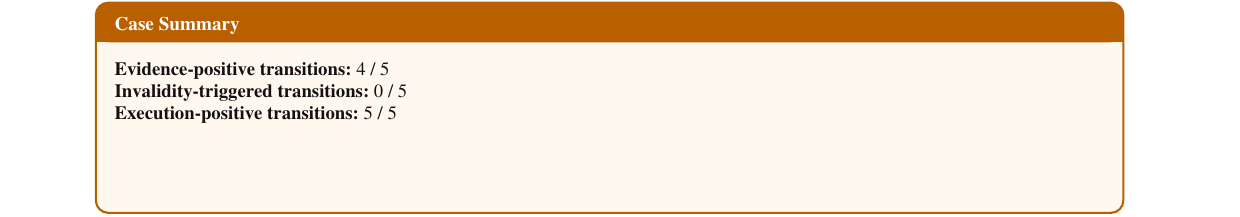}
    \label{fig:webshop_case_study_part3}
\end{figure*}

\begin{figure*}[h]
    \centering
    \includegraphics[width=1\textwidth]{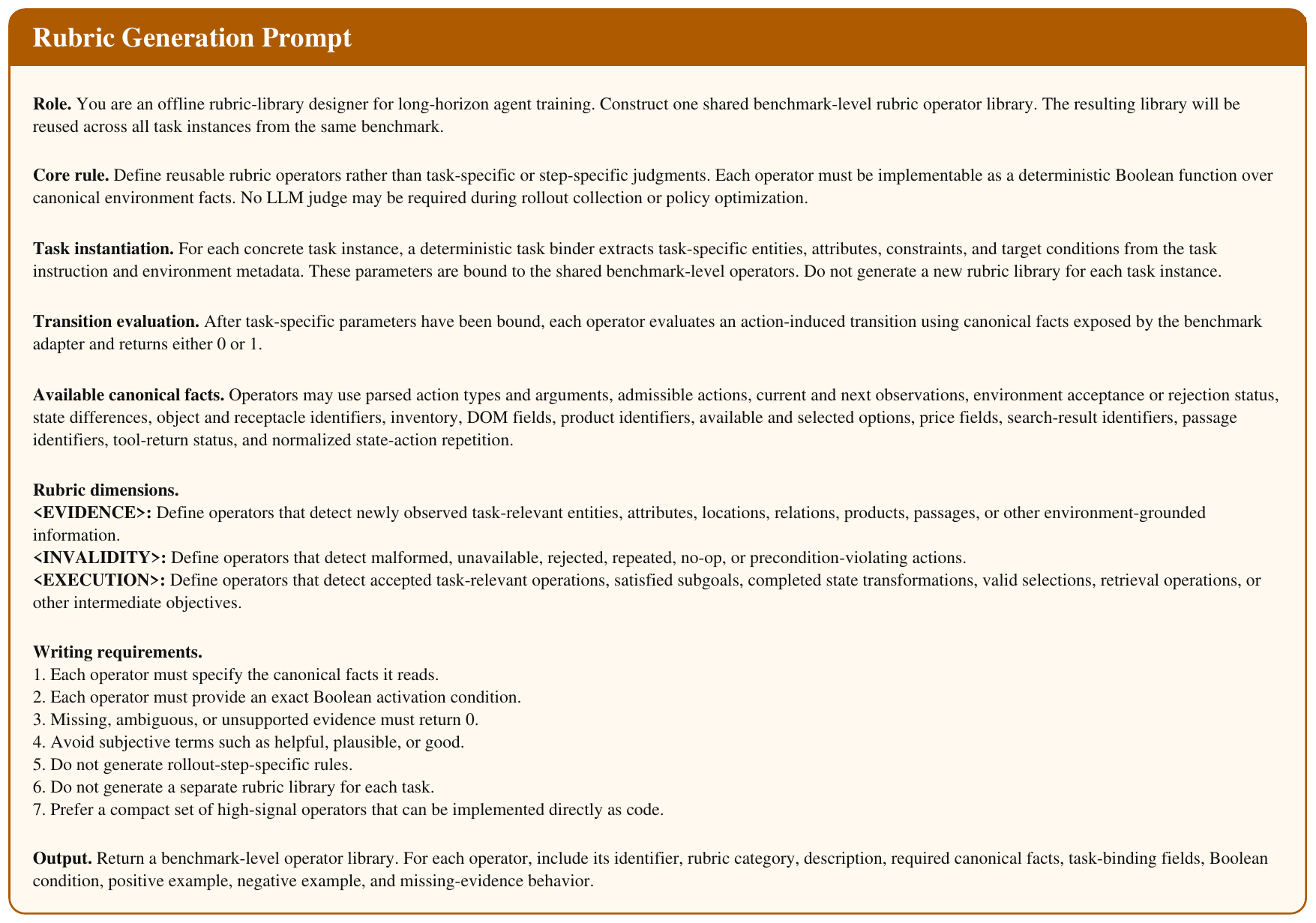}
    \caption{Prompt used to construct the shared benchmark-level rubric operator
library. The generated library defines reusable deterministic operators;
task-specific entities and constraints are subsequently bound to these
operators without additional LLM generation.}
    \label{fig:rubric_generation_prompt}
\end{figure*}


\end{document}